# From Mimicry to True Intelligence (TI) – A Nature-Inspired Paradigm for Artificial General Intelligence

Meltem Subasioglu[1,2,**] and Nevzat Subasioglu[3]

1. University of Konstanz, Department of Computer and Information Science, Konstanz, Germany
2. Google, Zurich, Switzerland
3. University of Konstanz, Department of Psychology, Konstanz, Germany

**Corresponding author. E-mail: meltem@subasioglu.ch; ORCID: 0009-0001-0171-2937

Abstract

The debate around Artificial General Intelligence (AGI) remains open due to two fundamentally different goals: replicating human-level performance versus replicating human-like cognitive processes. We argue that current performance-based definitions are inadequate, as they offer no clear roadmap for research and fail to define the qualitative nature of genuine intelligence. Four decades of work on cognitive architectures have already characterized the mechanisms; our claim is that this knowledge is not being consulted by the research program now driving AGI development, and what we offer is a translation of it into terms that bear on contemporary systems, together with criteria by which those systems can be assessed. Drawing on Cognitive Science, Natural Computing and the principles of Complex Adaptive Systems (CAS), we define True Intelligence (TI) as a system characterized by six core components: five architectural pillars for which assessment criteria can be stated – embodied sensory fusion, core directives, dynamic schemata creation, a highly interconnected multi-expert architecture, and an orchestration layer – together with Interconnectedness, an emergent quality for which we can offer no independent measure. Two commitments distinguish our framework from the architectures on which it builds: that the topology of the schema network is itself shaped by the directives under which a system has operated, so that identical experience under different directives yields differently organized knowledge; and that a continuously active associative substrate, from which a salience mechanism promotes only a fraction to deliberate processing, constitutes a distinct locus of cognitive work rather than the unconscious support of a task cycle. Based on the implementation of the five assessable pillars, we propose a five-level taxonomy of AGI, which we apply to four contemporary and classical systems. We hold that Level 5 is the strongest architectural condition for TI that can presently be specified, and we leave open whether meeting it is sufficient. By bridging classical cognitive architectures with contemporary AI challenges, this framework offers a research agenda addressed to a community that currently assesses progress by benchmark performance.

# 1 Introduction: The AGI Rift and the Need for a Clear Definition

The desire to create machines with human-like or human-level intelligence has been a guiding principle for the AI community since its early days. Yet still today, the meaning of "intelligence" and "human-level AI" remains unclear and contested [1, 2, 3]. This semantic confusion has led to a fragmented research landscape, where different groups pursue incompatible visions under the same banner of Artificial General Intelligence (AGI) [4, 5], leading to growing confusion in academia, industry, and broader societal debates. This is not merely a semantic issue but a deep-seated, value-laden conflict about the nature and purpose of AI, reflecting a fundamental tension between engineering a high-performance tool and the scientific pursuit of understanding and recreating the human mind. The notion of intelligence resists definition for a reason that no framework removes. The term was borrowed from ordinary language, where it never had a settled meaning. Competing technical definitions may therefore be talking past one another rather than disagreeing: not rival answers to a single question, but answers to different questions posed in the same word. We do not claim to resolve this. The definition offered below is a stipulation, put forward because it is architecturally consequential and assessable, not because we take it to capture what the word has always meant.

Aside from the lack of semantic clarity, two expressions are frequently used interchangeably in this debate. *Human-like* intelligence refers to thinking in the manner in which humans think, by means of comparable processes; *human-level* intelligence refers to thinking as well as humans think, that is, to comparable attainment. The two are logically independent. A system may reach human level by means quite unlike ours, and a system organized along human-like lines may perform well below human level. The framework developed in this paper is concerned with the former, and the failure to keep the two apart is another source of the confusion in this debate.

## 1.1 The Value-Laden Pursuit of "Human-Level" AI

As argued by [1], definitions of intelligence are what philosophers term "thick evaluative concepts", meaning they include both descriptive and normative elements, simultaneously describing empirical phenomena and assessing the desirability of specific behaviors. Consequently, any definition of AGI is not value-neutral; it embeds political, social, and ethical assumptions about what is worth building, by whom, and for what purpose. This value-laden character is a primary source of the fractured discourse, as different research programs implicitly champion competing values. The risk, as Ruha Benjamin warns, is that a narrow definition of what even counts as technology or intelligence can stifle the ability to imagine more just and equitable technological futures [6]. The current landscape, dominated by definitions that prioritize measurable utility, is shaped by the pragmatic demands of industry and venture capital, which favor metrics that demonstrate a clear return on investment [3, 7, 8]. The vocabulary proposed here aims to provide the precision needed for a more reflective and effective critique of these underlying values and to create space for a complementary, science-driven agenda.

This observation applies reflexively. The term we introduce, True Intelligence, is itself evaluative: to call one kind of system truly intelligent is to imply that others are not, and the word carries an approbation that a purely descriptive vocabulary would avoid. We considered more neutral alternatives and retained this one because the contrast it draws with mimicry is the argumentative point of the paper, but the reader should treat the label as a flag planted in contested territory rather than as a neutral description. What we do claim to be non-evaluative are the five architectural conditions themselves. Whether a system maintains a shared representational space across modalities, or revises a single knowledge structure without retraining, is a question about its organization, and the answer does not depend on whether one approves of the result.

## 1.2 The Two Competing Visions: Internal Processes vs External Performance

The AGI discourse can be understood as a conflict between two competing visions: one centered on process and the other on performance. This division mirrors a classic philosophical tension between functionalism (what a system *does*) and phenomenology (what a system *is*).

The performance-centric vision defines intelligence by its external capabilities and observable outputs. This perspective is exemplified by some of the most influential benchmarks and charters in the field. The Turing Test, a foundational benchmark, judges intelligence based on a machine's ability to generate responses that are indistinguishable from a human's, explicitly bracketing any concern for its internal cognitive state [4]. More contemporary examples follow a similar logic. OpenAI's charter defines AGI in terms of its ability to "outperform humans at most economically valuable work", a definition rooted in utility and economic value [5]. A recent and highly influential framework by Google DeepMind proposes "Levels of AGI" based on performance and generality, from "Emerging" (unskilled human) to "Superhuman" (outperforming all humans) [2]. This approach, while providing a good initial metric for performance, fundamentally focuses on *what* a system can do rather than *how* it does it, assuming the mechanisms are secondary. We argue that these definitions lack clear guidance of what AGI truly is, as measures of economic value or hypothetical benchmarks against human populations do not provide a clear directive on what research and engineering efforts should focus on to achieve it – the definitions are inherently empty of the architectural qualities that define genuine intelligence.

Two recent developments have sharpened rather than resolved this problem. First, the economic framing has become contractual: the agreement between OpenAI and Microsoft reportedly defines AGI as the point at which OpenAI's systems generate profits exceeding one hundred billion dollars [9], severing the term from any claim about capability or mechanism whatsoever. Second, the performance-centric position has been stated in its strongest form to date: writing in Nature, Chen and colleagues argue that frontier language models already satisfy reasonable standards for general intelligence, assessed not by a single test but through a cascade of increasingly demanding performance tiers [10]. Notably, they concede that we lack a detailed account of how the abilities of these models emerged, while maintaining that performance rather than mechanism should settle the question. We take that concession seriously and draw the opposite conclusion from it: a field that cannot say how the capabilities of its systems arise has no principled basis on which to direct its next decade of research. There is a second assumption in that argument, and it is easier to overlook than the first. An evidence cascade of this kind inherits the validity of the tests from which it is assembled: it can establish general intelligence only if each tier measures the capacity it is taken to measure. That assumption is not idle, and a substantial literature disputes it. Benchmarks presented as general measures of progress have been shown to operate as narrow and historically contingent artefacts whose construct validity was never established, and which stand in for the capacity of interest rather than measuring it [11]. Where a benchmark is deliberately constructed to resist memorization and search, as ARC-AGI-2 was, systems that had saturated its predecessor score in the low single digits [12]. We do not conclude that benchmarks are worthless. We conclude something narrower and, for the present argument, sufficient: a cascade of them cannot carry a claim about general intelligence, because what the tiers aggregate is not one property held fixed across them.

There is a further argument against performance-based definitions which we regard as stronger than the one just given. Definitions of intelligence in the psychometric tradition are careful to separate what an agent knows from its capacity to acquire and deploy knowledge. Ignorance of a domain is not a deficit of intelligence: no one considers a competent adult unintelligent for being unable to answer questions about a subject never encountered. Performance benchmarks confound precisely this distinction. A system that fails a benchmark may lack the capacity the benchmark purports to measure, or may simply not have met the relevant material during training, and the score cannot distinguish these cases. Conversely, a high score may register nothing beyond the presence of that material in the training distribution. This is not a remediable imprecision in particular benchmarks but a structural property of assessing a system by what it can currently do: such a measurement returns the joint product of capacity and exposure. Test items are demonstrably present in the pretraining corpora of the systems they are used to assess, and the resulting inflation cannot be quantified after the fact, since the corpora are neither public nor fixed [13]. Exposure is therefore not a residual to be estimated away but an uncontrolled variable in every such measurement.

In contrast, the process-centric vision posits that genuine intelligence must involve the replication of human-like cognitive processes. This perspective, often voiced by cognitive scientists, neuroscientists, and a subset of AI researchers, emphasizes the necessity of internal states like consciousness, sentience in the narrow sense of a capacity for subjective experience, and genuine understanding, distinguishing intelligence from what could be mere sophisticated mimicry [14, 15, 16]. From this viewpoint, a system that passes the Turing Test or excels at economically valuable work might still be a "philosophical zombie" – a system that perfectly imitates intelligent behavior without any accompanying subjective experience [17]. A classic illustration of this distinction is John Searle's Chinese Room argument, which challenges the idea that a machine can have a "mind" simply by following a program [14]. The thought experiment highlights that external behavioral competence, while impressive, does not necessarily equate to internal understanding or higher-level reasoning.

This line of argument requires care. The zombie objection has force only under the assumption that intelligence entails phenomenal consciousness, and that assumption is neither uncontroversial nor necessary for the argument developed here. Definitions pitched at what Newell termed the knowledge level, for instance, characterize intelligence in terms of the knowledge an agent has and the goals it pursues, abstracting over the mechanisms that process that knowledge without invoking experience at all [18]. We therefore do not rest our case on the zombie argument, and the framework below is constructed so that none of its claims depend on whether a system satisfying it has subjective experience. Our reason for attending to internal organization is different, and we take it to be stronger: a definition ranging only over external performance provides no guidance about what to build.

Process-centric work has also not been silent about *what* to build: the Newell test [19], the capability analyses of Langley, Laird and Rogers [20], and Goertzel's competency areas for AGI [21] each specify criteria a candidate architecture should satisfy. A separate difficulty attaches to those process-centric arguments that rest instead on notions such as sentience and consciousness, which cannot be operationalized or measured directly.

The consequence thus is not that this work is unavailable but that it reaches the wrong audience. Two qualifications are needed at once, since a stronger claim would be false. Work restating this tradition in terms of contemporary systems does exist and is recent: CoALA organizes language agents around modular memory, a structured action space and a decision cycle [22], and Wray, Kirk and Laird catalogue cognitive design patterns recurring across cognitive and agent architectures, using them to diagnose what present-day agents built on language models lack [23]. Section 3 discusses both in detail. Our observation is narrower than a claim of absence. That work is addressed to researchers who build agent architectures and asks how such systems should be constructed; it does not take up the question of what the field should be aiming at, nor does it supply criteria against which a claim to have reached general intelligence could be assessed. Such criteria remain formulated in the vocabulary of cognitive architecture and evaluated against the systems that discipline builds – while the claims about general intelligence that now set the field's priorities are made on the basis of benchmark performance. Supplying a restatement addressed to that second audience is among the aims of this paper.

Two terms need clarification before we proceed. First, we use sentience throughout in its narrow sense, as the capacity for subjective experience, including affective states such as pleasure and distress. Consciousness, as used here, is the broader notion encompassing both such experience and the functional access relations that make information available for reasoning and report; sentience in this sense is one component of consciousness rather than a synonym for it. Neither notion is measurable by the criteria we set out below, and our framework is deliberately constructed so as not to depend on either.

Second, about the term mimicry, which appears in our title and recurs throughout. Used without care it begs the question at issue: to describe a system's behavior as mimicry is to presuppose that nothing beyond imitation is occurring, which is a substantial part of what this paper undertakes to establish. We use the term in a narrower and, we hope, defensible sense. A system mimics a capacity when it reproduces the observable signature of that capacity by means that do not include the mechanism the capacity is ordinarily taken to require. On this reading the word carries no implication about experience, and whether a given system is mimicking is settled by examining its architecture rather than by intuition about what it is like to be that system. It follows that mimicry in our sense is not a term of disparagement, and that a system could mimic a capacity while being, on any reasonable measure, more useful than one that does not.

### 1.3 Bridging the Gap via Natural Computing

To resolve the tension between process- and performance-centric approaches, we find it productive to look beyond purely computational definitions and to treat intelligence first as a biological phenomenon, in order to ask which of its principles are substrate-independent enough to be realized on silicon. The field of Natural Computing [24] provides a crucial bridge, positing that genuine intelligence arises from specific principles of Complex Adaptive Systems (CAS) – such as self-organization, evolutionary adaptation, and feedback loops [24, 25].

In this paper, we propose a practical framework that bridges these two viewpoints. We leverage the principles of Natural Intelligence (NI) not as a requirement for “wetware”, but as an architectural blueprint for True Intelligence (TI) on silicon.

We thus want to highlight that this paper does not propose a new cognitive architecture. Decades of work in that tradition - from ACT-R [26, 27] and Soar [28, 29] to the Common Model of Cognition [30] - have produced a substantial body of mechanism-level knowledge about how integrated cognition can be organized, much of it grounded in behavioral and neural data. Our concern is that this knowledge has become effectively invisible to the research program that currently drives AGI development, where progress is defined by performance on benchmarks and, increasingly, by economic milestones. The gap is not one of evidence but of translation: what these mechanisms mean in engineering terms, how they map onto contemporary AI systems, where those systems fall short of them, and what research agenda follows. That translation is what we offer here, providing a Perspective: it argues for a direction of research and specifies criteria by which progress along it could be assessed, and it reports no implemented system. That limitation is real, and we would not present the framework as established on the strength of argument alone. What we have tried to provide in place of an implementation is a set of assessment criteria concrete enough to be applied to systems we did not build, and specific enough that a system could fail them. The final column of Table 3 states these for each pillar, and Section 4.1 applies them to four existing systems.

Following this introduction, Section 2 deconstructs human intelligence through the lens of foundational cognitive models and Complex Adaptive Systems. Section 3 relates the resulting picture to existing cognitive architectures. In Section 4, we leverage these insights to define the emergent pillars of True Intelligence (TI) and propose a five-level AGI taxonomy based on their implementation. Section 4.2 outlines a research roadmap that addresses the current gaps in building genuinely intelligent systems. Section 5 then addresses the philosophical "hard problem of consciousness" through a functional and architectural lens. Finally, Section 6 provides a concluding summary of our framework and its implications for the future of AI.

## 2 Deconstructing Human Intelligence: A Blueprint for Genuine Cognition

To define the path to True Intelligence (TI), we must first formalize what constitutes human-like intelligence beyond task completion. Human cognition is not a monolithic capacity, but a complex, multi-layered architecture built upon foundational cognitive models and giving rise to emergent properties. These components, drawn from decades of research in cognitive science and neuroscience, serve as the blueprint for our proposed notion of TI. Taken together, the subsections that follow are not a survey of cognitive science but a selection: each isolates one property of human cognition that is well established empirically, that is architectural rather than incidental, and that current AI systems either lack or approximate only superficially. They build on one another, from the foundational models and their neural substrate, through the distinction between fluid and crystallized capacity, to the schema dynamics, motivational grounding and orchestration that the later pillars formalize.

## 2.1 Foundational Cognitive Models

The classical view of human uniqueness, commonly traced to Aristotle although the emphasis on rationality as the defining human characteristic is more properly Kantian, often describes humans as "rational animals" [31, 32]. More modern perspectives highlight the "cognitive niche" humans occupy, leveraging their minds to outsmart competitors [33], or their "cultural niche", which allows knowledge accumulation across individuals and generations, enabling survival in diverse environments [34, 35, 36]. Regardless of the specific lens, the underlying mechanisms of human intelligence are multifaceted. According to various converging definitions, intelligence involves reasoning, problem-solving, and learning [37, 38, 39, 40]. Gottfredson further elaborates on this, emphasizing planning, abstract thinking, comprehending complex ideas, quick learning, and learning from experience [40]. These capabilities are often seen as emergent properties of anatomically distinct cognitive systems, each with its own capacity.

In this context, foundational models provide a crucial understanding of how these complex abilities are structured and function. David Marr [16], Roger Shepard [41], and Dominic Massaro and colleagues [42] all proposed a productive strategy for understanding human cognition: by analyzing the abstract computational problem underlying a cognitive function and using its ideal solution to understand human behavior. For Marr, this was the “computational level” of analysis [16]; for Shepard, it was a way to identify universal laws of cognition [41]; and, in the tradition of rational analysis to which their work contributed, a component of “rational analysis” [42]. This approach has since been applied to a wide range of problems, including reasoning [43], generalization [41], categorization [44], and causal learning [45].

Marr's levels of analysis have had a durable influence well beyond vision, supplying much of the vocabulary in which computational cognitive science still separates what a system computes from how it does so; the framework remains widely taught and applied, while also attracting sustained criticism for underweighting the constraints that implementation places on the levels above it. We adopt the distinction in the spirit in which it is generally used, as an analytical convenience rather than a claim of strict independence between the levels.

The collective outcome of all this work was the establishment of a paradigm that views cognition as a set of a-priori principles and computational problems to be solved, thereby providing a clear, mechanism-focused agenda for the study of intelligence. This mechanistic and computational perspective, which seeks to understand the "how" behind intelligent behavior, serves as a foundational pillar for our own framework.

The pursuit of modeling these foundational mechanisms has historical precedents in cognitive architectures like ACT-R [26] and Soar [28], which pioneered the modular replication of human thought. More contemporary proposals, such as Global Workspace Theory [46] or Yann LeCun’s Joint Embedding Predictive Architecture [47], have further attempted to map cognitive routing and world-modeling onto machine learning. However, classical architectures, notwithstanding substantial work on robotic embodiment in Soar and on perceptual-motor timing in ACT-R, largely developed their accounts of higher cognition apart from sensory grounding, while modern proposals frequently remain tethered to the optimization of static datasets. Our True Intelligence (TI) framework builds upon this lineage, directly contrasting these structural approaches against the 'crystallized intelligence' trap of current frontier models to highlight what is still missing.

## 2.2 Global Workspace Theory as a Cognitive Blueprint

Among contemporary cognitive theories, Global Workspace Theory (GWT) provides the most explicit architectural bridge between cognitive psychology and the mechanisms proposed in this paper. Originally formulated by Bernard Baars [46] and later extended into a neuronal Global Workspace model by Dehaene and colleagues [15, 48], GWT posits that the mind contains a vast ensemble of specialized, parallel, and largely unconscious processors – each expert in its own domain, from visual feature extraction to motor planning. Conscious cognition arises when a subset of these processors gains access to a capacity-limited "global workspace," where the selected content is then broadcast back to the entire network of specialized modules. In the theatre metaphor of Baars, this broadcast is what makes information globally available: once admitted to the workspace, a representation can inform planning, memory encoding, verbal report, and voluntary action.

For our framework, GWT is significant for three reasons. First, it provides principled grounding for the Orchestration Layer developed below, recasting it not as a routing heuristic but as a capacity-limited integrative bottleneck that resembles the role neuroscience assigns to the prefrontal-parietal network [48, 49]. Second, GWT makes the explicit prediction that general-purpose cognition requires such integrative broadcasting, and language models satisfy that prediction only in an attenuated form. The currently dominant frontier architecture is now a mixture of experts, in which a learned router selects among specialized sub-networks rather than passing every token through a single dense computation [50], and interpretability work has found functional specialization emerging within trained networks even where none was imposed [51]. Differentiation, which is one half of what the theory requires, is therefore present. What is absent is the other half. The experts of such a model share a single representational space and are selected per token by a router, so that there is no capacity-limited workspace to which one content is admitted, and no broadcast of that content back to modules holding separate representational commitments. Specialization without a workspace is not what GWT describes. Third, GWT is explicitly mechanistic without committing to biological implementation details, which makes it a natural target for artificial instantiation. Nevertheless, GWT describes a function rather than a recipe: it specifies what the architecture must do (admit a selected representation to a capacity-limited workspace and broadcast it back to the specialized modules) but does not prescribe how to build the selection, admission, and broadcast dynamics in silico. Our Orchestration Layer takes this unfinished blueprint and couples it to the adjacent pillars: the Dynamic Schemata Module provides the representational content to be broadcast, the Multi-Expert Architecture supplies the modules that consume the broadcast, and the Core Directives and Dynamic Schemata influence which contents win the competition for workspace access.

It should be noted that this is not a gap the cognitive architecture literature left unaddressed. The working memory of Soar and the buffer system of ACT-R already instantiate, in concrete computational form, what Global Workspace Theory characterizes abstractly: a store whose contents are globally available to otherwise encapsulated modules, with access to it governed by mechanisms these architectures specify precisely, in ACT-R through capacity-limited buffers and in Soar through a decision cycle that imposes seriality on an otherwise parallel substrate. What follows below is therefore not offered as the first computational reading of the theory, but as an attempt to say which neuroanatomical division of labor that reading corresponds to, and what follows for artificial systems if it is taken seriously. Section 3 returns to this literature in detail.

More than a decade ago, researchers utilizing resting-state neuroimaging identified three large-scale brain networks whose interactions govern the transition between unconscious and conscious processing – directly mapping to what the GWT describes functionally as the underlying neuroanatomical substrate [52, 53, 54]. The Central Executive Network (CEN), anchored in the dorsolateral prefrontal cortex and posterior parietal cortex, implements goal-directed, deliberate cognition: working memory, cognitive flexibility, and top-down attentional control [49, 55]. This is the neural substrate of the Orchestration Layer. The Default Mode Network (DMN), spanning the medial prefrontal cortex, posterior cingulate cortex, and hippocampus, is active during internally directed thought: autobiographical memory retrieval, prospective simulation, self-referential processing, and, crucially, the spontaneous and associative activation of schemata [52, 53]. The Salience Network (SN), centered on the anterior insula and anterior cingulate cortex, acts as the dynamic gatekeeper between the two: it detects biologically and cognitively significant signals and coordinates the suppression of DMN activity in favor of CEN engagement, or vice versa [54].

This three-network architecture provides the biological mechanism that the GWT specifies functionally. The DMN is the neural backbone of what GWT calls the unconscious specialist processors: the schema networks whose associative dynamics run continuously and in parallel, below the threshold of the global workspace. The SN is the biological correlate of the GWT's selection process, the mechanism by which sufficiently salient content crosses the threshold and gains access to the capacity-limited workspace. The CEN then operates on this admitted content, implementing deliberate, goal-directed reasoning. Critically, the Orchestration Layer in our TI framework corresponds not to a single region but to the CEN-SN complex: the CEN executes executive control once content is admitted, while the SN governs what is admitted in the first place.

It should be noted that the triple-network organization is not a proposal of ours. It was characterized in systems neuroscience over the past two decades and has been replicated across resting-state and task-based studies, with the Salience Network's switching role between the other two among its more robust findings [56, 57]. Its identification as the substrate underlying the phenomena that Global Workspace Theory describes functionally is, by contrast, an interpretation, and it is not the only available one: accounts situating these networks along a continuous gradient of macroscale cortical organization [58], or parcellations resolving the cortex into a considerably larger number of finer-grained networks [59], describe the same data differently and do not privilege a tripartite division. We adopt the tripartite formulation because it is the coarsest level of description at which architectural consequences can still be stated, not because we regard the boundaries between these networks as sharp or their number as settled.

An important finding with direct implications for our framework concerns individual variation in the functional relationship between the DMN and CEN. These networks were initially characterized as strictly anti-correlated, such that when one is active the other is suppressed [60]; that characterization has since been substantially refined [61]. For tasks that are intrinsically self-referential, such as prospective memory, theory of mind, and metacognitive monitoring, the DMN and CEN co-activate rather than compete [61]. Individuals with higher metacognitive ability, working memory capacity, and fluid intelligence exhibit greater flexible coupling between the two networks: they are not more capable because one network permanently dominates, but because they can suppress the DMN more precisely when focused external attention is required, and integrate DMN-sourced schema content into executive processing more effectively when the task demands it [62, 63]. Metacognition, by its nature, always requires this integration: monitoring one's own reasoning (a CEN function) about one's own schemas and memories (a DMN function) cannot be accomplished by either network in isolation. This individual-level variation is instructive for the TI framework. It reveals that Interconnectedness, the sixth and emergent pillar of TI, is not merely a structural property but one that can vary in degree. At the biological level, it is precisely the flexible, context-sensitive co-activation of the DMN and CEN orchestrated by the SN, rather than their mere co-existence, that differentiates a system capable of genuine metacognitive self-awareness from one that merely alternates between processing modes.

### 2.3 Fluid and Crystallized Intelligence: The Capacity for Novelty and Experience

In 1963, Psychologist Raymond Cattell first proposed that general intelligence is not a single entity but is composed of two primary factors: crystallized intelligence (Gc) and fluid intelligence (Gf) [64]. Gc refers to the ability to use knowledge, skills, and experience acquired over a lifetime; it is the application of what one has already learned [40]. Gf, in contrast, is the capacity to reason, solve novel problems, and identify patterns, independent of any prior specific knowledge [40].

These two factors are sometimes identified with the two systems of dual-process theory, and thereby with the distinction between deliberate and automatic processing that cognitive architectures draw. The identification is tempting and we do not endorse it. Gc and Gf are factors recovered from the covariance structure of test scores in psychometric research, and in the Cattell-Horn-Carroll synthesis they stand alongside a substantial number of further broad abilities [65]; they describe how individual differences in performance are organized, not by what processes that performance is produced. System 1 and System 2, by contrast, propose a distinction between processes. Nothing prevents a person high in Gf from being so by means of extensive automatic processing, and the correlational structure would not register the difference. We retain the Gc/Gf distinction because it names the specific asymmetry we observe in current systems, namely command of accumulated knowledge alongside brittleness on genuinely novel problems, and because instruments for measuring it exist. We do not treat it as a claim about mechanism. Section 2.5 takes up the process question separately, and Section 3 sets out what the cognitive architecture literature has already established about both.

While Gc tends to increase or remain stable throughout adulthood, Gf is known to peak in early adulthood and then gradually decline [66]. This distinction is critical for evaluating modern AI. Current frontier AI systems, particularly large language models (LLMs), trained on vast corpora of text and data, demonstrate a powerful command of crystallized intelligence. They can retrieve facts, synthesize information, and apply learned patterns with superhuman proficiency. However, their capabilities in fluid intelligence are far more brittle. This limitation is starkly revealed by benchmarks like the Abstraction and Reasoning Corpus (ARC), introduced by François Chollet to specifically measure abstract reasoning and novel problem-solving abilities [67]. ARC tasks are designed to be easy for humans, relying on innate cognitive priors [68], but exceedingly difficult for AI systems that rely on pattern matching from training data [67].

Despite the exponential scaling of LLMs between 2020 and early 2024, the state-of-the-art scores on the original ARC benchmark hovered around 34%, well below the estimated human baseline of over 60-70% [69]. While recent breakthroughs have pushed some systems' performance to surpass human scores on this benchmark, these results were often achieved using computationally intensive search and test-time adaptation methods, which highlight a reliance on computational power rather than a fundamental leap in cognitive architecture [69]. This struggle is further underscored by the introduction of the new, more challenging ARC-AGI-2 benchmark in 2025, which is specifically designed to resist these methods [12]. Early results show that even frontier AI models that excelled on the original benchmark struggle with ARC-AGI-2, scoring in the low single-digit percentages while humans can solve every task with high efficiency [12]. This persistent gap demonstrates that current AGI architectures excel at leveraging existing knowledge (Gc) but struggle with the genuine abstraction and efficiency required for novel problem-solving (Gf). The development of fluid reasoning in humans is neurobiologically grounded in a distributed fronto-parietal network, with the rostrolateral prefrontal cortex (RLPFC) playing a crucial role in relational integration – the ability to consider multiple relations simultaneously to solve a problem [70]. A system capable of TI must possess this capacity for fluid reasoning, moving beyond the mere application of learned knowledge to generate truly novel solutions [12, 67, 70].

Charles Spearman famously observed more than a century ago that human performance is correlated across a spectrum of seemingly unrelated tasks [71]. He proposed a dominant general factor "g" to account for correlations in performance between all cognitive tasks, with residual differences reflecting task-specific factors. Theoretical accounts such as the aforementioned Global Workspace Theory [15, 46] and Integrated Information Theory [72] posit that consciousness may have evolved as an architectural platform for exactly this kind of general-purpose integration: a mechanism that broadcasts salient information across specialized modules and coordinates their outputs into coherent, goal-directed behavior.

Recent psychometric work on AI systems provides an unexpected empirical test of this claim. Treating benchmark batteries as cognitive test batteries and model releases as subjects, Krakauer applied principal component analysis to 39 models evaluated on 14 benchmarks between 2019 and 2025 [73]. A positive manifold does emerge: through 2023 and early 2024 a single component accounted for as much as 92% of the variance in performance, the signature of a general factor. That factor then declines, to roughly 64% on the same battery, as reasoning-specialized models appear; effective dimensionality rises, and stable trade-offs become visible beneath the general component, such that gains on reasoning benchmarks come at the expense of execution benchmarks. The interpretation offered there is that capability dimensions are diversifying rather than collapsing onto a single axis. For the present framework the implication is direct: scaling has produced increasing specialization, not increasing generality. If general intelligence is, as Spearman's observation suggests, a property that emerges from integration across specialized capacities, then a trajectory along which those capacities become measurably less integrated is not a trajectory towards it.

## 2.4 Dynamic Schemata and Human Randomness

Drawing from the foundational work of Jean Piaget [74, 75] and Frederic Bartlett [76], schema theory posits that human knowledge is not a flat database of facts but is organized into dynamic, interconnected mental frameworks, or schemata [75, 76]. These schemata – representing concepts, events, and relationships – provide a structure for understanding the world and guide our expectations and inferences [77, 78]. They are continuously updated through two primary processes:

- **Assimilation:** New information is integrated into an existing schema without changing it [75, 79].
- **Accommodation:** An existing schema is restructured, or a new one is created, to account for new information that contradicts the current framework [75, 79].

This dynamic process allows for robust generalization and learning from every experience. To make this concept more precise, we can provide a preliminary mathematical formalization. We can model a schema $S_k$ as a probabilistic construct, defined by a set of parameters $\theta_k$. When a new piece of information $X_t$ is encountered at time $t$, the system assesses its relevance by computing an activation score for each schema. Using a Bayesian framework, the activation of schema $S_k$ can be represented as the posterior probability $P(S_k|X_t)$:

$$P(S_k|X_t) \propto P(X_t|S_k)P(S_k) \tag{1}$$

where $P(X_t|S_k)$ is the likelihood of observing the information given the schema, and $P(S_k)$ is the prior probability of that schema.

The decision to assimilate or accommodate is determined by this activation score.

- **Assimilation**: If a schema's activation score is above a predefined threshold $\tau_A$ (i.e., $P(S_k|X_t) > \tau_A$), the new information is assimilated. This process involves updating the schema's parameters incrementally without altering its core structure. This can be conceptualized with a simple update rule:

  $$\theta_{k,new} = (1-\alpha)\theta_{k,old} + \alpha * Update(\theta_{k,old}, X_t) \tag{2}$$

  where $\alpha$ is a learning rate and $Update$ is a function that modifies the schema's parameters based on the new data $X_t$.

- **Accommodation**: If no existing schema's activation reaches the assimilation threshold $\tau_A$, the system must accommodate the new information. This necessitates a more fundamental change. If a different, lower threshold $\tau_C$ is met for a related schema, it might be restructured. If the information is entirely novel, a new schema is created:

  $$\text{create } S_{new} \text{ with parameters } \theta_{new} \text{ from } Xt, \text{ if } \forall \text{k}, \text{P}(S_k|X_t) < \tau_C \tag{3}$$

This formalization provides a technical lens through which to view a system capable of dynamically updating its world model, a significant departure from the static, implicit knowledge of most current AI models. Work in cognitive science indicates that this updating process is stochastic rather than deterministic [80, 81]. We deliberately avoid the term non-algorithmic here: a stochastic process is not thereby a non-computable one, since randomized algorithms are algorithms, and the stronger reading is no part of our claim. This so-called "Human Randomness" is not a flaw but a constitutive feature of a Complex Adaptive System [24, 25]. Three observations make its role clear. First, stochasticity in cognition enables exploration and the breaking of local optima in uncertain environments – behavior that deterministic, gradient-based optimization cannot readily replicate without carefully engineered entropy bonuses. Second, human decision noise is neither independent nor uniform; it is context-sensitive and structured, often approximating the statistics of the environment rather than an idealized random process [81]. This structured noise appears central to fast, adaptive reasoning under uncertainty, and an AI system that systematically suppresses it in favor of maximum-likelihood outputs inherits the brittleness of its training distribution. Third, stochasticity is present across biological scales, from synaptic transmission through neural population variability up to the collective dynamics of behavior [82], and the Natural Computing literature argues that variability of this kind is exploited rather than merely tolerated: to regulate exploration, to prevent premature convergence, and to sustain long-term adaptability [24, 25]. Unlike static weights in a deep neural network, biological schemata therefore enable non-linear, context-aware adaptation, allowing the system to respond to novel inputs in ways that pre-programmed or statistically regularized algorithms cannot. Characterizing and reproducing Human Randomness is, accordingly, not merely a cosmetic refinement but a central empirical question for AI systems that aim for genuine general adaptability.

The practical challenge of building a system that can autonomously and robustly perform assimilation and accommodation on a complex, interconnected schema network remains a major open research problem. Our goal here is to clarify these foundational cognitive processes, not to propose a specific, novel algorithm to solve them.

A further architectural property of the schemata network with direct consequences for our TI framework is its internal topology. Within a single schema cluster, a set of representations united by shared context such as a specific episode, an emotional state, or a conceptual domain, connections are dense and recurrent, enabling partial cues to trigger full pattern completion without deliberate effort. This organization mirrors the CA3 region of the hippocampus, where recurrent collateral connections allow a fragmentary input (a scent, a sound, a visual fragment) to cascade through an entire associative memory chain through direct schema-to-schema activation [83, 84]. This is why associative chains of the kind described above – smell to childhood memory to emotion to abstract thought – unfold subconsciously: they propagate within the dense intra-schema connectivity of the DMN without requiring orchestrator mediation.

Between schema networks and the global workspace, however, the connectivity follows a fundamentally different pattern. The projections from specialized schema clusters to the broader cortical hierarchy are sparse and compressed, conveying a low-dimensional summary of the network's internal state rather than its full richness. This organization reflects the small-world topology observed throughout the cortex: dense local clustering within specialized regions, sparse long-range projections between regions [85, 86]. A small number of high-coverage long-range projections therefore suffices to propagate information globally, even though each conveys a compressed rather than a complete summary of its source. The consequence is architecturally significant: the Orchestration Layer does not have, and does not require, access to the full internal state of each schema network. It operates on salience-filtered, compressed summaries. Much of what the system is computing is therefore invisible to deliberate metacognitive analysis, not because it is inaccessible in principle, but because the topology does not route it through the global workspace by default. A TI architecture must therefore not conflate the Orchestration Layer's output with the totality of the system's cognitive activity: the system is computing far more than the orchestrator is even aware of. Section 3 records that ACT-R made this observation precise two decades ago, and states what we take to be new in the version of it developed here.

## 2.5 System 1 and System 2 Thinking

The dual-process distinction between fast, intuitive System 1 thinking and slow, deliberate System 2 thinking [80] is a useful descriptive metaphor but does not constitute a foundational architectural model. The three-network framework described above provides its mechanistic basis: System 1-like processing corresponds to DMN-driven schema activation running below the Salience Network's promotion threshold; System 2-like processing corresponds to CEN-driven deliberate cognition with active SN gating. Rather than treating these as discrete modes, the three-network model reveals them as endpoints of a spectrum governed by the SN's dynamic allocation between the DMN and CEN. Most current AI systems are, by this characterization, powerful System 1 engines: they excel at high-dimensional pattern matching. It was once possible to add that they lack the architecture for autonomous System 2 reasoning, and that techniques such as Chain-of-Thought [87] and Tree-of-Thoughts [88] elicit deliberation only through external prompting. That characterization no longer holds. Models trained by reinforcement learning to produce extended reasoning traces generate multi-step deliberation without any external scaffold, as a learned disposition rather than as a response to instruction [89]. The distinction we need is therefore not between prompted and unprompted deliberation, which no longer separates such systems from biological ones, but between deliberation elicited by a task and a substrate that runs whether or not a task is present. A reasoning-trained model deliberates when queried; between queries it does not continue to activate and recombine its representations, and no salience mechanism decides which of those activations warrant deliberate processing. That is the distinction the present taxonomy is designed to make precise, and Section 3 examines how Soar and CLARION treat the underlying question architecturally, which is one reason we do not rest our argument on the dual-process vocabulary.

## 2.6 Core Directives and Intrinsic Motivation: The Engine of Autonomous Exploration

Motivation in biological systems is not a monolithic concept but a structured system of basic drives from which more specific goals are subsequently derived [90]. This idea is well-established in fields like psychology and biology. For instance, Drive Theory posits that physiological needs create an aroused state that motivates an organism to satisfy them, while Maslow’s Hierarchy of Needs organizes human motivation from basic physiological and safety needs to higher-order drives like social connection and self-actualization [90, 91]. At the base of our framework, we establish what we call Core Directives, which are the most fundamental, hard-coded survival principles. In humans, this includes self-preservation, resource acquisition, and social connection – principles that correspond to the basic drives these classic theories describe and are wired within our genetic code. For an artificial system, core directives could be framed as foundational imperatives like maintaining computational integrity, ensuring energy supply, or fulfilling a base-level programmed purpose. These directives serve as the ultimate "why" behind all behavior, creating the high-level goals for a homeostatic system, significantly shaping how schemata are being created. The success and safety of such an AGI, however, depend critically on the choice of these directives and the degree to which they are aligned with human values, a non-trivial challenge that forms a central part of the AI alignment problem [92].

A further qualification is needed here. The strict ordering Maslow proposed, in which lower needs must be satisfied before higher ones become operative, has found only limited empirical support. MicroPsi accordingly treats needs as explicitly non-hierarchical: they act concurrently, and priorities among them are established by an individual weight attached to each need rather than by sequential unlocking [93]. Nothing in the framework developed here depends on a strict ordering. What it requires is only that a small set of basic directives grounds the goals that develop from them, and that those directives are not themselves acquired, a commitment MicroPsi shares, since there cognition can change goals, expectations of reward and priorities but cannot directly influence the needs themselves [93].

Arising from these directives is intrinsic motivation, which is the emergent drive to learn, explore, and master new skills for their own sake. According to Deci and Ryan's Self-Determination Theory, this motivation stems from the fulfillment of three basic psychological needs: autonomy (the feeling of volition), competence (the feeling of mastery), and relatedness (the feeling of connection to others) [94]. This internal drive fuels curiosity, exploration, and the open-ended acquisition of knowledge and skills that are essential for developing general intelligence. An AI system that is intrinsically motivated is not simply executing a programmed task; it is actively seeking out novel experiences that help it refine its internal world model, thereby better serving its underlying core directives.

In a reinforcement learning context [95], this can be formally expressed by a total reward function that combines both extrinsic and intrinsic rewards.

$$R_{total}(s_t, a_t) = R_{ext}(s_t, a_t) + \beta R_{int}(s_t, a_t) \tag{4}$$

Here, $s_t$ represents the state of the agent at time $t$, and $a_t$ represents the action taken by the agent at time $t$. $R_{ext}$ is the reward provided by the environment or a human designer (e.g., scoring points in a game), while $R_{int}$ is the internal reward signal that stems from the system's core directives. The coefficient $\beta$ controls the weight given to intrinsic motivation. A system driven by TI would have a significant, non-zero $\beta$, allowing it to explore and learn even in the absence of external rewards. A common but limited approach to intrinsic reward is to formulate curiosity as the prediction error of the agent's world model [96]:

$$R_{int}(s_t, a_t) = \|f(s_t, a_t) - s_{t+1}\| \quad (5)$$

While curiosity-based formulations of $R_{int}$ are a promising starting point, they remain externally engineered objectives that can be exploited by an agent focusing on irreducible environmental noise [97]. In contrast, evolutionary algorithms operate on a population of potential solutions, such as neural network weights, represented as a genome $\theta$ [25, 98, 99]. They optimize for a single, high-level fitness function $F(\theta)$, which evaluates the performance of a complete solution after it has finished a task or lifetime, without requiring a step-by-step reward signal. The optimization process over generations can be summarized as:

$$\theta_{t+1} = Evolve(Population_t, F) \quad (6)$$

This approach, similar to biological evolution, can lead to the emergence of complex behaviors that serve the system's core directives without the need for hand-crafted, sequential rewards. A TI system must possess a genuine internal drive for learning and competence that is not dependent on a hand-crafted reward signal, allowing it to set its own goals and engage in open-ended, self-directed development, as this intrinsic process is what drives the continuous assimilation and accommodation of new experiences into a dynamic schema network.

### 2.7 A Highly Interconnected Multi-Expert Architecture, the Orchestration Layer and Metacognition

Just as an intrinsically motivated system requires a purpose to guide its exploration, it also needs a cognitive architecture capable of effectively processing and integrating the resulting sensory and cognitive inputs. As noted previously, from a neuroscientific perspective, human intelligence is not a single, monolithic capacity, but rather a system of specialized, anatomically distinct cognitive modules. Each module functions as an expert in its own domain, such as vision, language, auditory processing, or motor control. The true power of this system, however, lies not in the individual experts but in their highly interconnected architecture. This foundational structure enables the deep, multi-directional flow of information that allows for a seamless and fluid fusion of diverse inputs, a capability essential for true multimodal processing and integrated problem-solving.

This multi-expert architecture is coordinated by a central executive function, which we define as the Orchestration Layer. This layer is functionally analogous to the human prefrontal cortex (PFC), which acts as the brain's central hub for managing information flow and decision-making [100], and corresponds to the Central Executive Network (CEN) highlighted above. Just like the PFC, the orchestration layer is responsible for integrating diverse inputs from the specialized expert modules, coordinating their activity, and enabling higher-order cognition [55, 101]. This cognitive control is not a simple routing mechanism but a sophisticated process that flexibly allocates cognitive resources and manages the system's workflow to achieve coherent, goal-directed behavior. The importance of this function is underscored by research indicating that the global connectivity of the PFC is a strong predictor of cognitive control and intelligence itself [49].

The synergy between the multi-expert architecture and its orchestrating layer gives rise to a crucial emergent capability: metacognition. Defined as "cognition about cognition", metacognition is the autonomous, internal ability to monitor, evaluate, and refine one's own thought processes [102, 103, 104, 105]. It allows a system to self-identify its own high probability of error, assess its confidence in a conclusion, and self-correct its logic without external prompting or supervision. This is a profound departure from prominent AI models that, while capable of complex tasks, often exhibit a lack of this self-awareness, confidently producing incorrect answers and failing to recognize their own limitations [106, 107]. The development of a robust orchestration layer that enables metacognition is therefore a critical milestone on the path to TI, as it allows a system to move beyond sophisticated mimicry and demonstrate a true, self-aware form of reasoning. We should say at once that architectural metacognition is not an unaddressed problem: Soar, CLARION and LIDA all implement mechanisms of this kind, and Section 3 sets out what they establish and where we believe something is nonetheless still missing.

### 2.8 The Imperative of Embodied Experience: The Bridge to Genuine World Understanding

The human brain's sophisticated cognitive architecture and orchestrating layer, while essential, do not operate in a vacuum. To ground abstract knowledge and enable true intelligence, a system must possess Embodied Sensory Fusion – the ability to ingest and process a vast stream of multimodal sensory inputs from the real world (or an advanced-enough virtual environment). This is a profound shift from disembodied data processing to an active, physical engagement with the environment [108, 109]. A physical body and real-world interaction are foundational because they provide the rich, intuitive understanding of physical properties, causality, and the system's position within its environment [110]. Without this grounding, abstract concepts remain disconnected from the physical reality they represent. A disembodied AI, for instance, might be able to generate a flawless textual description of the concept of "heavy" by analyzing countless sentences and data points. However, it cannot truly understand why a large, hollow object is easier to lift than a small, dense one, because it lacks the physical experience of force, weight, and effort. Its knowledge is purely syntactic, not semantic. This leads to the system building incomplete schemata representations of the world it operates in. The fusion of diverse sensory inputs – such as vision, sound, and touch – creates a single, coherent representation of the environment, a process known as multi-sensor data fusion [111]. This process allows the system to build a deep, intuitive world model, thereby providing the necessary foundation for the more abstract, higher-level cognitive functions of the orchestration layer to operate upon. Without this foundational pillar, the most sophisticated cognitive architecture would be at best a powerful reasoning engine for a disembodied simulation, forever locked in mimicry rather than possessing true, grounded intelligence.

This should, however, not be taken to imply that embodied agents, human or artificial, possess complete schematic representations of their environment. Much of what humans know is acquired second-hand and remains partly syntactic: a child may be told not to touch the stove rather than learning by burning, and color can be described to someone who has never seen it. Human schemata are likewise partial, uneven in resolution, and frequently anchored in testimony rather than experience. The claim here is comparative rather than absolute. Direct sensorimotor engagement supplies a class of constraint, on force, resistance, and the consequences of action, that no amount of textual description supplies, and schemata formed without any such grounding are systematically impoverished in that respect.

This assessment is not ours alone. A recent workshop report on the relationship between neuroscience and artificial intelligence identifies the inability to interact with the physical world as the first of three fundamental capability gaps in current systems, observing that the internal representations of large language models were not built through an agent's own causal interaction with physical reality [112]. The same report notes that every major architectural advance in the history of AI - from perceptrons to convolutional networks to transformers - delivered capabilities its predecessors could not achieve, and that none of them emerged from scaling what came before.

## 2.9 The “Bow-Tie” Architecture of Cognition

To understand how these diverse components – from embodied sensory fusion to the orchestration layer – cohere into a robust system, we can look to the principles of biological systems theory. A universal trait of robust biological systems, ranging from the immune system to cellular signaling networks, is the organization into a "Bow-Tie" architecture [113, 114]. This architecture is characterized by a "fan-in" of diverse inputs that are compressed into a narrow "knot" or core, which then facilitates a "fan-out" to a wide variety of outputs. In the context of True Intelligence (TI), this framework explains the structural necessity of the Orchestration Layer.

- **The Fan-In:** This mostly corresponds to the **Embodied Sensory Fusion** and the **Multi-Expert Architecture**. The system must ingest highly variable, noisy, and high-dimensional data from the environment (vision, audio, somatic feedback) and specialized cognitive modules.

- **The Knot (The Core):** This is the **Orchestration Layer** (analogous to the Prefrontal Cortex and the CEN). Its function is to compress this vast array of inputs into a coherent, low-dimensional control signal or decision state. It acts as the "constraint that deconstrains," providing a universal protocol that allows diverse modules to communicate without needing a direct connection to every other module [113].

- **The Fan-Out:** This represents the system's ability to generate a limitless variety of adaptive behaviors, **Dynamic Schemata**, and novel solutions from the core protocols.

The evolutionary advantage of this architecture is systemic robustness. It allows the system to handle high variability and novel perturbations in the environment (the inputs) without catastrophic failure, provided the core protocols (the knot) remain intact. However, this architecture also highlights a critical vulnerability: while robust to external fluctuations, bow-tie systems are fragile to perturbations within the core itself [114]. This theoretical insight reinforces the critical importance of our Core Directives (Section 2.6); if the fundamental drives or the orchestration logic within the "knot" are corrupted, the entire system’s capacity for intelligent adaptation collapses. Thus, the Orchestration Layer is not merely a manager of resources, but the structural guardian of the system's integrity against the entropy of an open-ended environment.

However, as a complete description of biological cognitive architecture, the Bow-Tie model requires an important distinction. Its defining metaphor, a single compulsory knot through which all information flows, implies that the Orchestration Layer maintains visibility over and causal control of all cognitive activity. The three-network model outlined above reveals why this is not the case: a substantial proportion of schema activation and associative processing unfolds within the Default Mode Network (DMN), entirely outside the Central Executive Network's (CEN) attentional scope, unless the Salience Network (SN) elects to promote specific content. Such a chain propagates through the dense intra-schema connections of the DMN without orchestrator mediation; it enters deliberate awareness only if the SN determines that a signal within that chain is sufficiently salient to warrant CEN engagement. The biological knot in the Bow-Tie is therefore not the CEN alone, but the CEN-SN complex operating as a selective gate over a continuously active and largely autonomous DMN substrate.

This qualification has a direct consequence for our TI framework. Any artificial system that routes all cognitive operations through a central orchestrator will fail to replicate the parallel, largely subconscious associative dynamics that characterize biological intelligence. A genuine TI system requires not only an Orchestration Layer but also a parallel schema-network substrate with its own continuous internal dynamics, whose outputs are selectively promoted to orchestrator attention by a salience-gating mechanism, not exhaustively processed by it. The Bow-Tie correctly captures the global topology, a fan-in of diverse inputs, a core protocol, and a fan-out of adaptive behaviors. What it cannot capture is that much of the most cognitively significant activity occurs within the dense recurrent connections of the schema networks before any information reaches the knot, and that it reaches the knot only when the SN decides it should. Section 3 records that combining a central bottleneck with substantial parallelism is an established commitment of the Common Model of Cognition rather than a proposal of ours, and that our contribution here is confined to the character of the substrate beneath the gate.

## 3 Relation to Existing Cognitive Architectures

The framework developed above draws on a research tradition that has pursued integrated models of cognition for four decades. That tradition is not a competitor to what we propose; it is the source of most of what we take as settled. Our argument is directed at a different audience, namely the research program currently driving AGI development, which has largely ceased to consult it. Before identifying what we believe is still missing, we therefore set out what this literature has already established, because the gaps become visible only against that background. Comprehensive surveys of the field are available [20, 115]; here we restrict ourselves to the architectures bearing most directly on the five pillars; Table 1 summarizes the resulting attribution.

**Soar** [28, 29] organizes behavior around a working memory whose contents are tested in parallel by procedural rules, with a single operator selected per decision cycle. Where available knowledge is insufficient to select or apply an operator, or where no operator is proposed at all, Soar generates an impasse and automatically creates a substate in which the agent deliberates about the decision, about the application of the operator, or about the creation of a new one; the results are compiled back into new rules by chunking. Laird describes the resulting organization as three levels of processing above the architectural implementation: parallel activity within modules, the decision cycle itself, which corresponds to Newell's level of deliberate acts, and substate processing, which he characterizes as metalevel reasoning extending over many cycles. The parallel to System 1 and System 2 is considered explicitly in that account, where two candidate mappings are weighed and one is judged overly simplistic, with the conclusion that the categories resist mapping onto specific computational processes without further analysis. We share that caution, and it is one reason we do not rest our argument on the dual-process vocabulary. Soar is equally instructive about the limits of what has been achieved: metacognitive capability is reported there as partial and directed principally at reasoning about action, and while an experimental version has used appraisal-based intrinsic reward [116], no such reward belongs to the architecture as it stands. Reinforcement learning and episodic and semantic memory, by contrast, are established components, though semantic memory is populated deliberately rather than by an automatic learning mechanism.

**ACT-R** [27] shares the central production system but is distinguished by the specificity of its biological commitment. Its modules communicate through capacity-limited buffers, and both the buffers and the modules behind them are associated with identified neural structures: the goal buffer with dorsolateral prefrontal cortex, the retrieval buffer with ventrolateral prefrontal cortex, the long-term declarative store behind that buffer with the temporal lobes and hippocampus, and the matching, selection and execution of production rules with the striatal-pallidal-thalamic loop. The mapping is not illustrative. ACT-R models are tested directly against imaging data: given a commitment to an information-processing model and to the regions its buffers occupy, the architecture predicts the relative peak latencies and the relative areas under the BOLD curves across conditions without free parameters, the shape-fitting parameters being estimated per region. Predictions of this kind have been confirmed for the manual buffer in motor cortex, the imaginal buffer in posterior parietal cortex and the retrieval buffer in lateral prefrontal cortex. Its authors are equally explicit about where the account fails: dorsolateral activation was not observed in symbolic transformation tasks, the caudate response they report is one they could not derive from the theory, and they note that not all processing passes through the production system, since direct connections between cortical areas exist. Two further aspects deserve emphasis here. First, retrieval from declarative memory is governed by an activation equation combining a base-level term, learned from the history of use, with associative activation from the current context, so that contextual influence on retrieval is quantitative rather than set case by case. Second, ACT-R already articulates the observation that the central system does not have access to the full state of its modules, that people are aware of the object currently attended to rather than of the whole visual field, and of the fact currently retrieved rather than of the whole of memory. Our own treatment of compressed inter-module projections in Section 2.4 restates a principle this architecture made precise two decades ago.

**MicroPsi** [93, 117] develops the pillar that the architectures oriented towards task performance leave least specified: where goals come from. Motivation is not absent elsewhere. CLARION devotes a subsystem to it, and LIDA supplies it through feelings and emotions; but CLARION's own account describes its coverage of motivation as rudimentary [118], and LIDA's motivational apparatus belongs at present to the conceptual rather than the implemented model [119]. MicroPsi is where this pillar is worked out in most detail. Rather than presupposing a goal, it posits a minimal, orthogonal set of systemic needs, physiological, social and cognitive, which signal deviations from their set-points as urges. Goals are then learned as the situations and actions that have satisfied or frustrated those urges, and a motive is an urge bound to such a learned goal together with a plan for reaching it. The strength and urgency of active needs additionally modulate processing globally, setting arousal, the resolution level at which representations are processed, a suppression parameter defining the selection threshold that governs whether a new urge may interrupt current activity, and a securing rate determining the direction of attention between the environment and internally generated content. This is the most fully elaborated formalization of what we have termed Core Directives that we are aware of, and our treatment should be read as building upon it. It also corrects an assumption in our own presentation: needs in MicroPsi are explicitly non-hierarchical, acting concurrently with individual weights rather than in Maslovian succession, and we adopt that formulation here.

**CLARION** [118] contributes the most explicit dual-representation commitment: four subsystems, action-centered, non-action-centered, motivational and metacognitive, each divided into an explicit top level and an implicit bottom level that are engaged within the same processing cycle and whose outputs cooperate in both action selection and learning. The claim that a symbolic, deliberate process runs alongside a subsymbolic one, reactive in the action-centered subsystem and associative in the non-action-centered one, is thus already architectural rather than metaphorical.

CLARION also rejects the notion of a central general-purpose symbolic processor outright, which is a precedent worth noting for the discussion of the Orchestration Layer below.

**LIDA** [119] is the most direct precursor of the present framework: a systems-level model of cognition, partially implemented, in which a cognitive cycle moves repeatedly through understanding, attending and action selection. Attention is taken, following Baars, as the process by which content is brought to consciousness, and motivation is supplied by feelings and emotions rather than by built-in drives. Its authors distinguish carefully between the conceptual model and the implemented one, and the motivational apparatus belongs at present to the former; they also make no commitment to phenomenal consciousness, restricting their claims to functional consciousness. LIDA is grounded in cognitive science and neuroscience and has been positioned by its authors against six of the nine open problems set out by Langley, Laird and Rogers [20]. Any claim that the components we assemble have never been assembled before would fail against LIDA, and we make no such claim.

A consensus across part of this field has been formalized as the Common Model of Cognition [30], originally presented as a standard model of the mind. It began as a consensus reached at the 2013 AAAI Fall Symposium on Integrated Cognition and was extended through a synthesis across three architectures, ACT-R, Soar and Sigma, with CLARION, LIDA, MicroPsi and others named by its authors as candidates for later incorporation. The Common Model specifies that processing is parallel both within and across modules while a cognitive cycle selects a single deliberate act at intervals of roughly fifty milliseconds in human cognition, which supplies, in its authors' words, "the seriality necessary for coherent thought in the face of the rampant parallelism within and across components". Combining a central bottleneck with substantial parallelism is therefore an established commitment of this literature, and our discussion of the Bow-Tie topology in Section 2.9 should be read as describing that solution rather than proposing one. The translation of this body of work into contemporary AI has also begun. CoALA [22] both surveys existing language agents and offers a blueprint for building them, organizing both around modular memory, a structured action space and a decision cycle. Wray, Kirk and Laird [23] identify design patterns recurring across cognitive and agent architectures, among them observe-decide-act, commitment and reconsideration, and knowledge compilation, and propose one further pattern, step-wise reflection, as newly emerging from systems built on language models; they use these patterns both to diagnose what such systems currently lack and to derive specific empirical questions about them. The two differ in kind: CoALA is a blueprint for construction, whereas Wray and colleagues present cognitive design patterns as an analytic and comparative instrument, a companion to architectural implementation research. We share their premise that the goal of the field is to identify the mechanisms sufficient for general intelligence, and their criterion for assessing a system is architectural, namely whether and how completely it instantiates a recurring pattern. Our question is the prior one: what the target of AGI research should be, and by what criteria progress towards it should be assessed, at a point where the field answers both by benchmark performance.

| Pillar | Established in | What that work provides | What remains open for AGI systems |
|---|---|---|---|
| Embodied Sensory Fusion | ACT-R perceptual-motor modules; robotics-facing work in Soar | Perception as attention-limited encoding into buffers; timing-realistic perceptual-motor coupling | Fusion at the representational level, rather than modality adapters attached to a frozen model |
| Core Directives | MicroPsi; CLARION motivational subsystem; appraisal-based intrinsic reward in an experimental version of Soar | Needs as set-point deviations; goals learned from need satisfaction; global modulation by need strength | Directives shaping the organization of the world model, not only the selection of goals |
| Dynamic Schemata | ACT-R declarative memory and activation; Soar semantic and episodic memory; CLARION implicit level | Graded, context-sensitive retrieval; incremental online learning; implicit-explicit duality | A continuously active substrate whose associative dynamics proceed without mediation; structure-level transparency and targeted revision |
| Highly interconnected Multi-expert Architecture | ACT-R modules and buffers; CLARION's modular bottom level; Common Model of Cognition | Encapsulated specialists communicating through limited channels | Deep multidirectional transfer between specialists, rather than routing between frozen experts |
| Orchestration Layer | Soar impasses and chunking; LIDA global workspace; CLARION metacognitive subsystem | Metacognitive monitoring, reflection and compilation as architectural mechanisms | Deliberate control operating over a substrate it does not exhaustively supervise |

**Table 1:** Provenance of the five pillars. For each pillar, the cognitive architecture literature in which it is already established, what that work contributes, and what remains unresolved for contemporary AI systems.

Read together, this literature settles more than is generally acknowledged in current AGI discourse. What it does not settle falls into three areas, and these are the areas in which our framework makes its claims.

The first concerns what happens when no task is being performed. The architectures organized around task performance, ACT-R and Soar among them, structure cognition as a cycle in which procedural memory tests the contents of working memory and selects a single deliberate act that modifies it; the Common Model, which is a statement of consensus about what a cognitive architecture must contain rather than an architecture itself, generalizes this organization. Retrieval from declarative memory is cue-initiated, and long-term memory, perception and motor modules access only their associated working-memory buffers, so that a substrate of the kind we have in mind could not deliver content of its own accord; ACT-R's authors themselves note that direct connections between cortical areas exist and that their architecture may need to accommodate them. Several architectures come closer. CLARION's bottom level is consulted on every processing cycle rather than invoked on demand, though the subsystem holding its associative memory is, in its author's words, under the control of the action-centered subsystem, and its metacognitive subsystem monitors all the others. LIDA operates largely asynchronously and admits content to functional consciousness only through a salience-based competition, which is closer still. MicroPsi allows an urge to be satisfied opportunistically without significant attentional processing. What none of them develops is a substrate whose associative dynamics constitute a distinct locus of cognitive work in their own right, running independently of any current task rather than as the unconscious support of a perception-driven cycle. Yet the network whose activity defines the resting human brain is precisely the one active when no task is being performed, and the associative chains it sustains are among the more cognitively consequential things human minds do. This is a matter of specification rather than exclusion, and the distinction matters. The Common Model states that significant parallelism occurs within modules and that components, once initiated, run independently on their own time course, while leaving the details of their internal processing unspecified; such a substrate is therefore not ruled out by it. Its authors are explicit that omission from the model is often a statement of where a consensus is needed rather than a consensus on a lack of existence or importance, and we intend the observation in exactly that spirit: as an identification of where a consensus has yet to form, not as an attribution of oversight. For a system operating under persistent core directives, such a substrate would have to run alongside task processing rather than only in its absence, which is what makes the question consequential rather than academic.

That these omissions are recognized within the tradition itself is evident from current work extending the Common Model to metacognition [120] and to emotion [121]. Our reading of the situation is therefore not that this literature has overlooked the properties we emphasize, but that the research program now setting the agenda for AGI has engaged with neither its settled parts nor its open ones.

The second concerns the form knowledge takes. Declarative memory in these architectures is structured, inspectable and revisable; in the systems that currently define the state of the art it is none of these. Knowledge in a large language model is distributed across weights in a manner that makes a single fact impossible to remove without retraining, difficult to attribute to any locus, and opaque to inspection. A schema-structured representation is not merely a theoretical preference in this respect: it makes targeted revision, attribution of an output to the structures that produced it, and forgetting as a distinct operation tractable. These are among the problems on which current systems visibly fail, and this literature has held a candidate answer to them for decades.

The comparison with chunking deserves to be made explicit, since it is the closest existing mechanism. Chunking compiles the outcome of deliberate processing in a substate into rules that subsequently fire in parallel, so that what required reflection becomes automatic. What it does not do, and is not intended to do, is alter the organization of the knowledge over which it operates: the compiled rules enter an existing procedural memory whose structure the architecture fixes. The claim we make about Dynamic Schemata is of a different kind, and correspondingly harder to satisfy: that the clustering of representations, their internal density and the sparseness of their projections to the rest of the system are themselves shaped by the directives under which the system has operated, so that two systems with identical experience but different directives would organize the same knowledge differently. We are not aware of an architecture that makes this claim, and we regard it as the point at which our framework is most clearly falsifiable.

The third concerns integration itself. The consensus model specifies modules communicating through a shared working memory. It does not specify how a system whose specialists are learned rather than engineered, as every capable contemporary model's are, should be coupled such that transfer between them is deep rather than nominal. Attaching a vision encoder to a frozen language model yields a system with two modalities and one representational space, and this is not what the architectures above describe. That criticism, however, no longer reaches the strongest current systems, which are trained across modalities end to end and do achieve a genuinely shared representation; the objection has to be restated rather than repeated. What remains absent is not shared representation but differentiated specialists between which information is transferred. A jointly trained multimodal model has one representational space and no modules, which is the opposite failure from the adapter-based systems and equally distant from what this literature describes. The term multimodal, as currently used, covers both.

Our position relative to one assumption of the consensus model should be stated plainly. The Common Model commits to the claim that no additional specialized architectural modules are necessary for complex cognitive activities such as planning, language processing and theory of mind, while explicitly allowing that architectural primitives specific to those activities can be included [30]. The commitment is thus to non-necessity rather than to prohibition, and Sigma's realization of several capabilities above rather than within the architecture is recorded there as a partial rather than a substantive disagreement. The Orchestration Layer described here is naturally read as such a module, and the burden this creates is correspondingly specific: an architecture positing a dedicated executive must explain why the function cannot be left to emerge from sequences of cycles, and must avoid reintroducing the homunculus that the consensus position was formulated to exclude. Our answer is that the layer is not an inspector standing above cognition but a gate. It does not supervise the associative substrate; it receives what a salience mechanism promotes from it, and it has no access to the remainder. Whether this is best realized as a distinct architectural module or as a regularity over cycles is an empirical question we leave open. What the framework requires is the functional separation between a substrate that runs continuously and a deliberate process operating on a selected fraction of it.

None of these observations is a criticism of the architectures discussed. They are, rather, the reason we regard that tradition as the appropriate foundation for AGI research, and its neglect as costly. A research program that defines its goal by benchmark performance, and increasingly by commercial milestones, has no place to put any of this. The taxonomy proposed in the next section is intended as one way of making these commitments legible to that program: not a new architecture, but a statement of which architectural properties a system must possess before claims about general intelligence are warranted, and in what order they can reasonably be pursued.

Since the preceding discussion has conceded a great deal, it is worth stating plainly what remains. Two claims in this paper are not, to our knowledge, made by the architectures above, and they are the two on which the framework should be judged. The first is directive-shaped schema topology: that the clustering of representations, their internal density and the sparseness of their projections are themselves shaped by the directives under which the system has operated, so that two systems with identical experience but different directives would organize the same knowledge differently. This is a claim about what determines representational structure, and it goes beyond the established finding that retrieval is context-sensitive and that compilation is experience-driven. The second is the continuously active, salience-gated substrate: that associative dynamics running independently of any current task constitute a distinct locus of cognitive work, from which a salience mechanism promotes only a fraction to deliberate processing, so that the system computes substantially more than its orchestrator can inspect. This is a claim about what a cognitive architecture must contain, and it is stronger than the uncontroversial observation that much processing is unconscious. Both are falsifiable, and the final column of Table 3 states for each what would count as evidence. We note also what follows if they are false: the framework would then reduce to a restatement of the Common Model in a vocabulary addressed to a different audience, which would still be worth doing, but would not be a contribution to cognitive architecture.

# 4 A New Taxonomy: From AGI to True Intelligence

As stated, human intelligence – and therefore, TI – is characterized by several emergent properties that are largely absent in current AGI systems. These pillars are not independent modules but deeply interconnected functions that, together, form the basis of genuine understanding and unparalleled adaptability. Their synergy is crucial; for instance, fluid reasoning requires System 2 deliberation, which is guided by metacognitive assessment, and the resulting knowledge is integrated into dynamic schemata. This process of schema refinement is, in turn, often driven by an intrinsic motivation to learn and master the environment and influenced by underlying core directives. A high-density sensory input stream, similar to human perception, enables a system to formulate a deeper understanding of its environment and its place within it, creating a virtuous cycle of learning and understanding.

While individual elements such as multi-agent routing or sensory fusion are actively researched, it is the specific synthesis of these components, for which biological cognition provides our only worked example – bound by intrinsic core directives and subjected to the structured stochasticity we have termed human randomness – that distinguishes True Intelligence from mere functional mimicry.

However, a qualification is required here. We do not claim that these properties are essentially biological, in the sense of being realizable only in or derivable only from a biological substrate. Cognitive architectures have produced comparable capabilities without appealing to biology at all, and the argument of this paper would not be weakened were a purely functional route to them established. The role we assign to biology is evidential rather than constitutive. The human brain remains the only system known to possess general intelligence, and its organization therefore constitutes evidence about which architectural commitments suffice and which combinations are mutually compatible. Where that evidence is specific enough to constrain design, as in the division of labor between the networks sustaining internally generated and externally directed processing, we treat it as a guide. Where it is not, we do not.

Given the nuanced understanding of human cognition outlined above, we can now provide a formal, assessable framework that bridges the gap between performance-based AGI and cognitively-grounded TI. Sections 2.1 to 2.9 and Section 3 identified properties of human cognition that are empirically established and architectural in character; the components below are those properties restated as engineering requirements, at the coarsest granularity at which they can still be assessed in an artificial system. Their number is not a claim about cognitive completeness. Nor is their order a ranking of importance: it reflects dependency, in that sensory grounding and directives condition what schemata form, schemata and specialized modules provide the content over which orchestration operates, and metacognition presupposes all of the preceding. Interconnectedness, the sixth and emergent property, is not represented here, and the reason should be stated precisely, since we described it in Section 2.2 as varying in degree and cited coupling measures in support. Both claims stand, and they are compatible. The biological case admits measurement because an independent anatomical description of the networks is available, so that coupling between them can be quantified without presupposing what the coupling produces. No such independent description exists for an artificial system whose modules are individuated by the very functional criteria we are trying to assess, and a measure of integration defined over components that were identified by their integration would be circular. Phi was proposed for exactly this purpose and is the natural candidate [122]; we do not adopt it because its computation is intractable for systems of the relevant scale and because its application to artificial architectures remains contested. Interconnectedness is therefore left unmeasured here for want of a non-circular procedure, not because we hold it to be unmeasurable in principle.

The relation of the five components to existing lists of required capabilities should be stated directly, since the overlap is substantial. The Newell test [19], the capability analysis of Langley, Laird and Rogers [20] and Goertzel's enumeration of competency areas [21] each specify what a candidate architecture must account for, and much of what they specify is presupposed here rather than restated: our Multi-expert Architecture and Orchestration Layer together cover what those lists treat as modularity, control and metacognition, while Dynamic Schemata covers memory, learning and generalization. Three differences are worth naming. First, those lists are addressed to researchers building architectures and are formulated in that vocabulary; ours is addressed to a community that assesses systems by benchmark performance, and each component is therefore paired with a statement of what would count as evidence that a given system possesses it. Second, they enumerate capabilities a mind must have, whereas we enumerate structural commitments a system must make, which is why our list is shorter and why its elements are not independent. Third, several capabilities those lists include, notably language, social cognition and development over the lifespan, are absent here. We regard them not as unimportant but as consequences of the five rather than as further requirements, and we note that this is a claim on which the framework could be shown to be wrong.

With this, we propose a five-level classification system for AGI, where a Level N AGI is defined by the successful implementation of N of the five components from our TI framework.

This approach moves beyond "mimicry" and provides a robust, actionable framework for evaluating progress towards TI. The specific challenges of measuring each component are detailed in our research roadmap in Section 4.2 and further elaborated in Section 6.1, where we outline the necessary benchmarks for each pillar.

## 4.1 Formalizing the AGI Levels

To evaluate both biologically inspired architectures and current ungrounded AI models (such as large language models), we define the AGI level by the number of components successfully implemented, but strictly governed by a set of architectural prerequisites. This constraint prevents illogical classifications – such as a system possessing an orchestration layer without underlying expert modules to orchestrate – while still allowing purely digital, disembodied systems to be mapped onto our taxonomy.

Let $C = \{C1, C2, C3, C4, C5\}$ be the set of five core, measurable architectural components of True Intelligence (TI), where:

- $C1 = \text{Embodied Sensory Fusion}$ $(the\ perceptual\ foundation)$
- $C2 = \text{Core Directives}$ $(the\ motivational\ driver)$
- $C3 = \text{Dynamic Schemata Module}$ $(the\ evolving\ world\ model)$
- $C4 = \text{Highly} - \text{interconnected Multi} -$ $\text{expert Architecture}$ $(specialized\ processing)$
- $C5 = \text{Orchestration Layer}$ $(central\ executive\ and\ metacognition)$

Let $C_S \subseteq C$ be the set of components successfully implemented by system $S$. A system's architectural configuration $C_S$ is considered *valid* if and only if it satisfies the following structural dependencies:

1. $C_5 \in C_S \Rightarrow \{C_4, C_3\} \subset C_S$
   *(An Orchestration Layer functionally requires a Multi-expert Architecture to manage, and Dynamic Schemata to inform its metacognitive routing).*

2. $C_4 \in C_S \Rightarrow (C_1 \in C_S \vee C_3 \in C_S)$
   *(A Multi-expert Architecture requires either embodied sensory streams or an internal schematic world model to provide cross-domain processing data).*

If system $S$ possesses a valid configuration $C_S$, its AGI level is defined by the cardinality of that set:

$$Level(S) = |C_S| \tag{7}$$

"Successful implementation", in this context, refers to a holistic evaluation of both structural and functional evidence:

**Structural Evidence** refers to the existence of a dedicated, identifiable architectural component within the system that corresponds to one of the five elements. This means the component is not merely a byproduct of a monolithic architecture but a distinct module with a defined purpose.

This requirement raises a question the framework must answer: how is structural evidence to be obtained in a system whose structure was learned rather than designed? For an architecture assembled module by module the question does not arise, since the modules are given by construction. For a trained network it is the entire difficulty, and the discipline addressing it is mechanistic interpretability, whose object is precisely the recovery of functional structure from learned parameters. Its findings bear on our criterion in both directions, and we take both seriously. In our favor, sparse autoencoders recover features from production-scale models that are interpretable, causally efficacious and abstract, including features that generalize across modalities [51]; functional differentiation is therefore present in trained systems and can be identified without having been designed in. Against us, the same work finds those features represented in superposition, distributed across neurons and pervasively polysemantic, which sits badly with our demand for a distinct module with a defined purpose. If specialization is real but not modular, our structural criterion is drawn at the wrong grain. We therefore restate it in the form the evidence supports: what is required is identifiable functional differentiation with a demonstrable causal role in the system's processing, established by intervention rather than by anatomical separability. A component counts as implemented if ablating or steering the features that realize it changes the behavior it is supposed to produce, and not otherwise. This also resolves an apparent inconsistency in our position, since the biological systems we take as evidence are themselves developmentally emergent rather than designed, and it would be incoherent to require of artificial systems a modularity that the biological case does not exhibit. The consequence is that structural evidence becomes an empirical claim about a system rather than an inspection of its blueprint, and it is why we regard interpretability research as a precondition for applying this framework to learned systems at all.

**Functional Evidence** refers to the observable behavior of the system that demonstrates the intended purpose of the component. This goes beyond traditional benchmarks by evaluating how a task is solved, not just whether it is solved. For instance, to demonstrate the successful implementation of the Orchestration Layer, we would look for evidence of dynamic task-routing and knowledge integration across different modules, rather than just a high score on a specific test.

Under this framework, a purely text-based agent enhanced with an updating knowledge graph and intrinsic motivation could achieve a Level 2 rating (e.g., implementing C3 and C2). This flexibility allows current AI research to be accurately located within the taxonomy. However, as argued previously, to reach the holistic synergy and true semantic understanding entailed by a Level 5 rating, the system must integrate the physical grounding of Embodied Sensory Fusion (C1). The dependency structure alone does not force C1 at Level 4, and we should be explicit about the tension this creates with Section 2.8, where we described a system without such grounding as locked in mimicry. Our position is that C1 is necessary for the full range of grounded understanding and is not necessary for the other four pillars to be instantiated, which is why a system can reach Level 4 without it while remaining incomplete in exactly the respect Section 2.8 identifies. What we do retract is any suggestion that the cardinality metric captures this. Counting implemented components treats the absence of C1 as one deficit among five, whereas we hold it to be the deficit that the remaining four cannot compensate. A Level 4 system lacking C1 is therefore not four fifths of the way to True Intelligence in any meaningful sense, and the number should not be read as implying that the pillars are interchangeable or that progress along the taxonomy is linear.

A framework of this kind is worth little if it is never applied, and the example just given is hypothetical. We therefore apply the criteria to four systems we did not build: a frontier multimodal language model of the reasoning-trained generation, a robotics foundation model of the RT-2 and π0 class [123, 124], a Soar agent equipped with episodic and semantic memory [125], and LIDA [119]. Two of these are contemporary and two belong to the tradition surveyed in Section 3, which is deliberate: a taxonomy that could only ever be applied to the systems its authors favor would be worth nothing. The judgements below rest on published architectural descriptions rather than on our own testing, and we have marked as partial every case in which the published record left us uncertain. They should be read as an illustration of how the criteria discriminate, and as an invitation to dispute particular cells, which the criteria are specific enough to permit.

| Component | Frontier multimodal LLM (reasoning-trained) | Robotics foundation model (RT-2 / π0 class) | Soar agent with episodic and semantic memory | LIDA |
|---|---|---|---|---|
| **C1 Embodied Sensory Fusion** | No. Natively multimodal models do maintain a shared representational space, but every modality is acquired from observed data; none is acquired through the system's own action, and the revised criterion is therefore not met. | Yes. Vision, language and action occupy a single representational space, and the action modality is acquired through the system's own interaction with an environment [123, 124]. | Partial. Perceptual-motor coupling and robotic embodiment have been realized, but fusion occurs symbolically, after modality-specific pre-processing. | Partial. Sensory memory and perceptual associative memory are specified in the model; embodied realizations remain limited [119]. |
| **C2 Core Directives** | No. The objective is supplied externally during training. The system holds no persisting needs and initiates no exploration in the absence of a prompt. | No. Behavior is conditioned on externally specified task instructions. | No. Appraisal-based intrinsic reward exists in an experimental version [116] but not in the architecture as it stands. | Partial. Motivation via feelings and emotions is specified, but belongs to the conceptual rather than the implemented model [119]. |
| **C3 Dynamic Schemata Module** | No. Knowledge is distributed across weights and a single structure cannot be revised without retraining. External retrieval changes what is available to the model, not how its knowledge is organized. | No. Inherits the representational properties of its language-model backbone. | Yes. Semantic and episodic memory permit targeted retrieval and revision, and chunking compiles deliberate processing into rules [125]. Our further requirement, that topology be directive-shaped, is not met. | Yes. Perceptual associative, transient episodic and declarative memories are updated within every cognitive cycle [119]. |
| **C4 Highly interconnected Multi-expert Architecture** | Partial. Mixture-of-experts routing selects among specialists within one representational space [50], but bidirectional influence between specialists is not demonstrated. | Partial. As for the language-model backbone, with an action head attached. | Yes. Distinct modules with defined interfaces communicate through a shared working memory. | Yes. Specialized codelets operate in parallel and compete for admission to the workspace. |
| **C5 Orchestration Layer** | Partial. Reasoning-trained models generate multi-step deliberation without external prompting [89], which the earlier characterization of this class understated; but no architecturally distinct executive monitors the process, and unprompted self-correction remains unreliable [106]. | No. | Yes. Impasses generate substates in which the agent deliberates about its own decisions, characterized by its authors as metalevel reasoning [28, 29]. | Yes. Attention codelets govern admission to a global workspace and thereby what is deliberately processed [119]. |
| **Resulting level** | **1 at most** | **1 to 2** | **3** | **3** |

**Table 2:** The five criteria applied to four systems. Judgements are based on published architectural descriptions, not on testing by the authors; partial marks a case in which the published record does not settle the question. The two classical architectures score higher than the two contemporary ones, which is the expected result of assessing architecture rather than performance, and is not a claim about their relative capability.

Three things follow from this exercise, and none of them is that the classical architectures are better systems. They are not: on any performance measure the contemporary models are incomparably more capable, and a Soar agent cannot do what a frontier model does. What the table shows is that capability and architectural completeness have come apart, which is the entire contention of this paper. The systems that satisfy more of our conditions are the ones that can do less, and this is precisely why a field that measures only the latter has stopped noticing the former. Second, the exercise identifies where the contemporary systems are weakest, and it is not where the discourse usually places them: C2 and C3, motivation and knowledge structure, are unmet by every contemporary system in the table, while the pillar most often discussed, multi-expert organization, is the one where they come closest. Third, no system in the table reaches Level 4, and none is close to Level 5, which should temper the claim that general intelligence has been achieved on any architectural reading of that term.

Our taxonomy rejects the idea of a simple performance hierarchy and instead provides a developmental roadmap. It also locates precisely where the questions of AGI and of TI meet, and where they come apart.

We contend that once a system has achieved a fifth-level AGI by implementing all five assessable components, the difference between it and True Intelligence (TI) is no longer one that architectural evidence can adjudicate. It is important to be precise about what follows from this, and about what does not. Level 5 is the strongest architectural condition we are able to specify: it collects those structural commitments for which we can state what would count as evidence, and a system meeting all five has satisfied every requirement this framework knows how to impose. Whether such a system thereby possesses True Intelligence is a further question, and the five components do not settle it, for the sixth – Interconnectedness – is precisely the one for which we can offer no independent measure. The consideration that would close the gap, namely that consciousness arises as an emergent byproduct of integrated, higher-order cognition [48, 72], is a position held within the literature rather than a result established by it, and we decline to rest the framework upon it. Our claim is therefore the weaker one: Level 5 is a necessary architectural condition for TI and the most demanding one we can currently formulate, so that a system failing it is not a candidate; whether it is also sufficient remains open. For a research program this is the more serviceable claim, since it specifies what must be built without requiring prior agreement about what will thereby have been achieved.

### 4.2 The Path to True Intelligence: Addressing Current Gaps and Future Directions

The journey from current AGI research to TI requires a paradigm shift away from simply scaling existing models and toward developing integrated architectures that explicitly target the core mechanisms of human cognition. This roadmap outlines five critical and deeply interconnected research areas. For example, a neurosymbolic architecture provides the means to combine intuitive and logical reasoning, which can be structured into specialized modules. These modules, in turn, can be coordinated by a global workspace, grounded through embodied experience, and driven by intrinsic motivation to continuously refine a schematic world model.

Table 3 provides an overview of this roadmap. An objection arises at this point which we concede in part. The criteria in the final column of Table 3 are behavioral: each states what a system must be observed to do. A paper contending that mechanism matters, and that performance measures are inadequate to it, might be expected to state its criteria mechanistically instead; the recommendation of ARC-AGI-2 in Section 6.1, which is a performance benchmark, sharpens the objection further. Our position is that behavioral criteria are admissible when they are diagnostic of structure rather than of capability, and that this is what separates them from the benchmarks we criticize. Targeted revision of a single knowledge structure without retraining is a behavior, but no system lacking structured, addressable knowledge can produce it; that is why we ask for it. A high benchmark score, by contrast, is compatible with almost any internal organization, which is precisely our complaint. Where we invoke ARC-AGI-2 it is as evidence about the persistence of the fluid-reasoning gap, not as a criterion for any pillar, and it should not be read as a test for Level 5. We accept that this leaves the framework one step removed from what it recommends: the ideal instrument would be a structural probe applied directly to the system, and Section 4.1 set out why we regard interpretability research as the route to one.

| Pillar | Goal | Why it's Necessary | Example Limitation | Promising Research Areas | How it is assessed |
|---|---|---|---|---|---|
| *Embodied Sensory Fusion* | Ingest and process vast multimodal sensory inputs from the real world (or sophisticated simulations) | Enables a rich, intuitive understanding of the physical world by grounding abstract concepts in continuous sensory experience. | A disembodied AI that generates a perfect description of "heavy" but cannot grasp why a large, hollow object is easier to lift than a small, dense one. | Embodied AI, Robotics, Multi-sensor Fusion [108, 109, 110, 111, 126, 127] | A shared representational space spanning at least two modalities, at least one of which is acquired through the system's own action in an environment. Evidenced by cross-modal transfer without retraining, and by sensitivity to the consequences of the system's own interventions rather than to observed correlations alone. |
| *Core Directives* | Fundamental survival drives from which higher-level intrinsic motivation emerges. | The foundational engine of autonomous, goal-directed learning and continuous evolution. | An AI that halts exploration when its extrinsic reward function is exhausted, lacking the internal drive to continue learning or set new, self-defined goals. | Evolutionary Algorithms, Novelty Search [25, 98, 128, 129] | Goal-directed exploration persisting in the absence of external reward, together with demonstrable derivation of subordinate goals from the declared directives. |
| *Dynamic Schemata Module* | Dynamically create, modify, and remove highly interconnected schemata to organize knowledge and experiences. | Enables robust generalization, contextual understanding, and adaptive behavior by continuously updating a world model from every experience. | An AI that has only learned to saw wood may fail to understand that a different material, like fabric, should not be sawed, demonstrating a lack of a unified schema of material properties and their response to force. | Neuro-symbolic AI, Cognitive Architectures, Online Learning [130, 131, 132, 133, 134] | Targeted revision of a single knowledge structure without full retraining, and application of an existing schema to a domain absent from training. For the directive-shaping claim: a demonstrable difference in how acquired knowledge is organized between two instances given identical experience under different directives. |

| Pillar | Goal | Why it's Necessary | Example Limitation | Promising Research Areas | How it is assessed |
|---|---|---|---|---|---|
| *Highly interconnected multi-expert Architecture* | Fuse and coordinate information from specialized modules with deep, multidirectional information transfer. | Mimics the brain's functional organization for true multimodal processing, flexible problem-solving, and coherent integration, extending narrow multimodal fusion approaches. | An AI that correctly identifies a video of a ball bouncing and the accompanying sound of a drum but fails to understand that the two events are causally related – for example, a musician is using the ball to hit the drum. | Multi-agent Systems, AI Orchestration Platforms [135, 136, 137] | Bidirectional influence between specialists, evidenced on tasks in which information held by one must constrain the processing of another. |
| *Orchestration Layer* | A central executive function that coordinates information flow and enables metacognition. | Integrates diverse inputs, coordinates specialized regions, and drives metacognitive processes to achieve coherent goal-directed behavior and self-correction. | An AI generating a detailed, but entirely fabricated, medical diagnosis based on an image, without flagging the absence of supporting evidence or the low confidence of its conclusion. | Metacognitive AI, Agentic AI, Self-monitoring Pipelines [88, 106, 138] | Unprompted self-correction: the system flags its own low confidence or error and revises its output without external instruction. For the substrate claim: evidence that representations continue to be activated and recombined while no task is being processed, and that only a salience-selected fraction of this activity reaches deliberate processing. |

**Table 3:** Roadmap for Achieving True Intelligence (TI)

### Embodied Sensory Fusion

To achieve genuine world understanding, an AI system must first be able to process sensory information with a richness and density that mimics a biological organism. This requires a paradigm shift from disembodied data processing to Embodied Sensory Fusion – the ability to ingest and process vast, multimodal data streams by actively interacting with the real world or a sophisticated simulation [108, 109, 110]. This is the foundational layer that enables a system to formulate a deeper understanding of its position in the world, its physical properties, and the causality of its actions. Current research in Embodied AI and Robotics [108, 109] is a promising path forward, as are advances in Multi-sensor Fusion that aim to create a single, coherent representation of the environment from diverse inputs like vision, sound, and touch [111, 126, 127].

### Core Directives

Core Directives (Section 2.6) raise a concrete engineering question: how can a system develop structured intrinsic motivation without relying on hand-crafted extrinsic reward signals? Several research traditions offer partial answers and are therefore promising for the development of this pillar. Evolutionary Algorithms do not require a detailed reward signal but instead optimize for a high-level fitness function over many generations, allowing complex, goal-directed behaviors to emerge without step-by-step supervision [25, 98]. Novelty Search methods push this further by explicitly rewarding the discovery of new solutions or behaviors, fostering open-ended, self-directed learning that naturally serves a system's core directives [128, 129]. Combining these evolutionary approaches with curiosity-style intrinsic reward formulations [96] offers a plausible route toward systems whose goal-directedness is structurally, rather than procedurally, specified.

### Dynamic Schemata Module

The human ability to adapt and learn from every experience is rooted in the continuous creation and refinement of schemata. A True Intelligence (TI) system requires a Dynamic Schemata Module that can dynamically create, modify, and remove these highly interconnected mental frameworks to organize knowledge and experiences. This is a key departure from current AI models. While both biological and artificial neural networks learn by adjusting weights in response to new information, a fundamental difference lies in how this knowledge is structured and organized.

In current AI models, knowledge is implicitly encoded in static weights, creating a massive but largely opaque black box. This makes it difficult to modify specific knowledge clusters. Whenever something needs to be adjusted or "forgotten", the entire model must be retrained, which is computationally intensive and inefficient. In contrast, the human brain, while its individual neurons may be "black boxy", organizes its knowledge into distinct, specialized neural clusters that process specific information, such as horizontal lines or a particular scent. These modules are then seamlessly interconnected to form a dynamic schemata network, allowing for complex, associative reasoning. This is possible because the network consists of highly interconnected schemata, which in turn store experiences with rich metadata – such as core feelings, temporal cues, and contextual links – enabling a more profound and adaptable form of reasoning. Research in Neurosymbolic AI is particularly promising for this, as it aims to bridge the gap between statistical pattern recognition and symbolic reasoning, allowing for the creation of a dynamically updating and explainable world model [133].
Cognitive Architectures and Online Learning frameworks also provide a fertile ground for developing systems that can continuously integrate and restructure knowledge as they encounter new data [130, 131, 132, 134].

### Highly interconnected Multi-expert Architecture

Just as the human brain is not a monolithic processor but a network of specialized, interconnected regions, a TI system requires a Highly interconnected Multi-expert Architecture. This architecture provides the structural foundation for true multimodal processing, allowing for deep and fluid information transfer between specialized modules – a stark contrast to current approaches that often rely on a stitched-together fusion of different modalities. Research in Multi-agent Systems and AI Orchestration Platforms is essential for this pillar, as it focuses on developing the tools and frameworks needed to build and manage networks of specialized, collaborating agents that can dynamically fuse information and solve complex problems in a coherent, integrated manner [135, 137].

### Orchestration Layer

At the pinnacle of this cognitive architecture is the Orchestration Layer, which acts as the central executive function, analogous to the human prefrontal cortex [49, 55, 100, 101]. This layer is the foundation for coordinating information flow, decision-making, and, most importantly, enabling metacognition. A truly intelligent system must possess an internal, autonomous ability to monitor, evaluate, and refine its own cognitive processes. Research into Metacognitive AI is dedicated to building agents with this self-monitoring capability [107]. Agentic AI and Self-monitoring Pipelines represent a promising path forward, as they seek to create autonomous systems that can self-correct their own logic, manage their own workflows, and achieve coherent, goal-directed behavior without constant external supervision [88, 106].

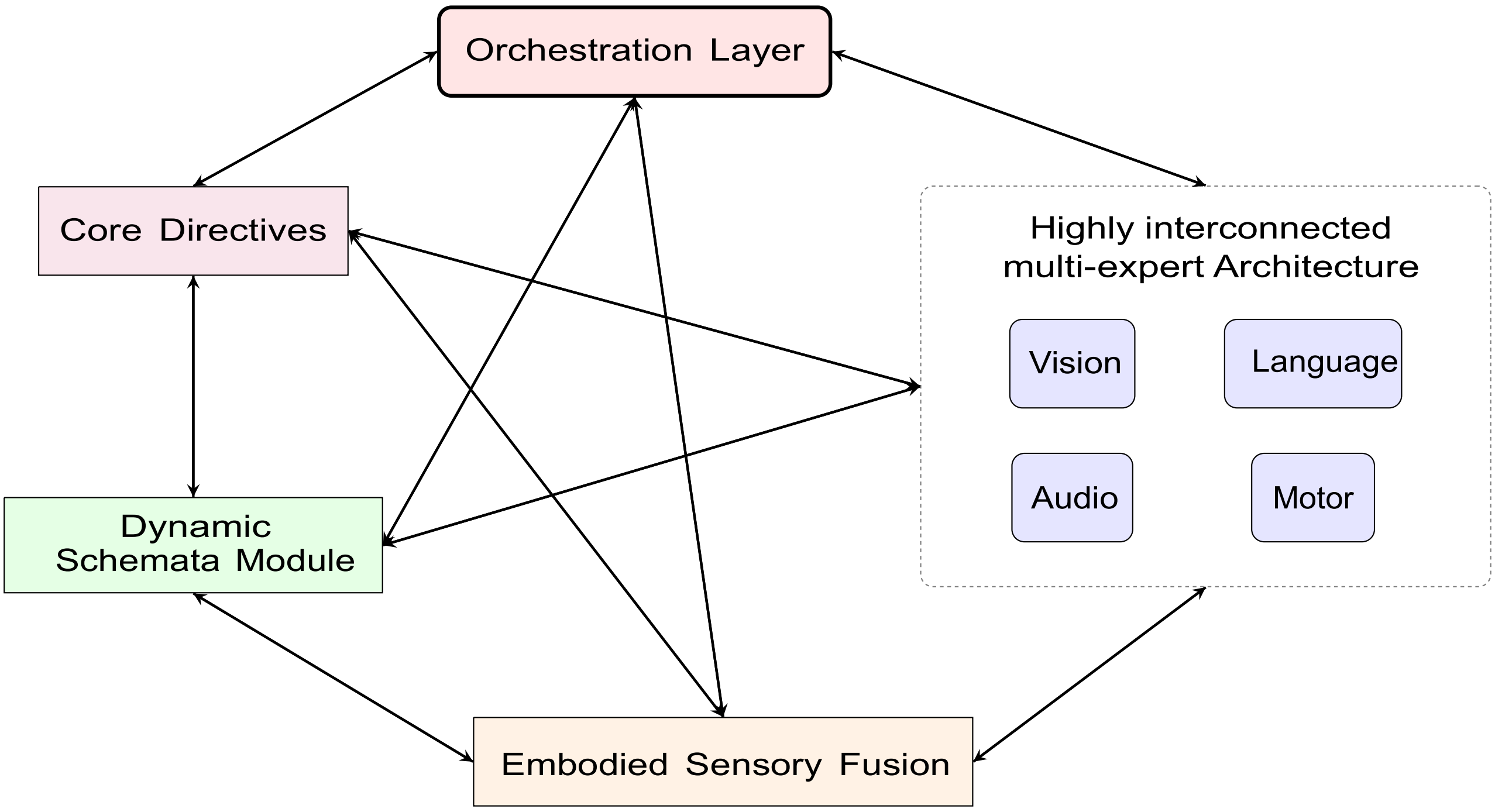

**Figure 1:** High-Level Architecture of True Intelligence (TI). The presented architecture outlines a system of five highly interconnected core components. The Orchestration Layer acts as the central hub for deliberate processing, coordinating the information admitted to it, while Core Directives provide the foundational motivation for all behavior. Embodied Sensory Fusion continuously feeds real-world data into the system, which is processed by a highly interconnected Multi-expert Architecture of specialized modules. The Dynamic Schemata Module serves as the evolving knowledge network, constantly updated by the system's experiences. This pervasive and bidirectional communication between all components is what we term Interconnectedness. The figure is a conceptual overview rather than an implementation diagram, and the connections it shows are the property the framework treats as emergent and, for the reasons given in Section 4, as resistant to independent measurement.

Figure 1 summarizes the framework as a whole and is best read as the culmination of the preceding analysis rather than as an implementation diagram. Three features carry the argument. The pillars are drawn as mutually connected rather than stacked, because none of them functions in isolation: directives shape which schemata form, schemata supply the content over which orchestration operates, and orchestration in turn determines which sensory and schematic content is integrated. The Orchestration Layer is central but not exhaustive, in keeping with the qualification developed in Section 2.9: it governs what enters deliberate processing without supervising the substrate from which that content is drawn. And Interconnectedness is depicted as a property of the connections rather than as a further box, which is precisely why it resists separate measurement and is treated in this framework as emergent rather than as a sixth implementable component.

# 5 The Hard Problem of Consciousness: A Functional Approach

This framework does not seek to solve the philosophical "hard problem of consciousness" directly but rather to provide a functional and architectural path toward its emergence. The "easy problems" of consciousness – explaining the functional aspects of the mind [139] – fall within the scope of our five AGI levels. The pursuit of TI aims to address the hard problem by creating the necessary conditions for conscious experience to arise. We argue that by implementing a system with all the measurable cognitive functions that in humans are inextricably linked to consciousness (metacognition, integrated information, self-driven learning, and a coherent sense of self-agency), we are, in effect, creating the conditions for qualia to exist. Qualia, in the sense used here, are the qualitative characters of experience: what it is like for a system to undergo a given state, as distinct from the functional role that state plays in its processing. This approach shifts the focus from a purely metaphysical question to a more tractable engineering and scientific challenge: to build a system whose functional organization satisfies the conditions that leading empirical theories associate with conscious experience, while remaining explicit that this leaves the further question open. Two further subsections follow the discussion of consciousness in this section, on the convergence of AGI and TI research and on the measurement problem. Their placement here is deliberate rather than incidental: both concern what remains once the architectural criteria have been stated, and the measurement problem in particular arises precisely because the sixth component resists the treatment the other five receive. A reader interested only in the taxonomy may take them with Section 4 without loss.

Two complementary empirical theories of consciousness – Integrated Information Theory and Global Workspace Theory – motivate this functional stance. Integrated Information Theory (IIT), anticipated in work with Edelman on the relation between consciousness and integration [72] and formulated by Tononi [122], proposes that consciousness corresponds to a system's capacity to integrate information in an irreducible manner, captured by the measure Tononi introduces as $\Phi$ (phi) [122]. On this view, any physical system with sufficiently high $\Phi$ is, to some degree, conscious – a strong, substrate-neutral claim that directly licenses the question of machine consciousness. For our taxonomy, the architectural implication is explicit: the high degree of integration observed in biological cognition is not a side-effect of a densely-connected substrate but a prerequisite for unified experience. A disembodied large language model, whose modules are instantiated as sequential forward passes through a single parameter matrix, achieves only weak integration across tokens and none across genuinely distinct modalities; its architecture therefore falls outside the class of systems to which IIT would ascribe meaningful $\Phi$.

The Global Workspace account (introduced in Section 2.2) complements IIT by offering a dynamical, functional criterion: consciousness is what happens when information enters and is broadcast from a capacity-limited central workspace [15, 46, 48]. On this account, consciousness is not a static property but a recurring ignition event – a transient, distributed synchronization that makes information available for report, recall, and deliberate control. Critically, both theories converge in rejecting the idea that raw computational power is sufficient: they point instead to specific architectural signatures – irreducible integration on the IIT side, workspace-style broadcasting on the GWT side – as the relevant empirical correlates of conscious experience.

The relationship between the framework and neural theories of consciousness has recently been approached from the direction of cognitive architecture as well: a mapping of four such theories onto the Common Model of Cognition finds that they jointly depend on recurrent local modules together with a cognitive cycle operating over a global working memory, and treats that architecture as a useful level of abstraction at which the theories can be compared and extended to artificial agents [140]. We read this as convergent with the position taken here. It also bears on the point at which our own account is least secure, since the connection between intelligence and phenomenal consciousness remains, on any current account including ours, considerably less concrete than the functional criteria that surround it.

Our framework is therefore deliberately agnostic about which theory – IIT, GWT, or their successors – will ultimately prove correct. It commits only to a weaker and more practical claim: any system that implements all five measurable TI pillars will also satisfy the functional and architectural prerequisites that both leading empirical theories of consciousness identify as necessary. Whether such a system is phenomenally conscious in the philosopher's sense may remain contested; whether it is functionally equivalent to a conscious agent would, at that point, no longer be a matter of dispute. This positions the hard problem as an asymptotic question rather than a prerequisite: we can make measurable, mechanism-based progress along all five pillars without first having to resolve what consciousness "really is."

## 5.1 The Convergence Hypothesis

While current pursuits for AGI and the proposed framework of TI can be distinct goals, the paths to achieving them are not entirely separate. The pursuit of a highly capable AGI, one that can perform across a vast array of tasks with human-level generality, will inevitably force researchers to confront and engineer solutions for many of the same problems central to TI. Building an AGI that can robustly handle truly novel situations and learn efficiently from limited data (requiring better generalization mechanisms like dynamic schemata), as well as operate across multiple modalities will demand the development of sophisticated architectures for memory, reasoning, and multi-modal processing.

These advanced functional components, developed in the pursuit of AGI, become the necessary building blocks for TI. AGI research, in this sense, provides the "body parts" – the powerful neural modules for vision, language, and planning – while TI research provides the "integrative blueprint" – the cognitive architecture that can weave these parts together to give rise to genuine understanding, intrinsic adaptability, and the conditions for consciousness. The two research paths are therefore symbiotic; progress in engineering high-performance systems provides the raw materials for the scientific quest to understand and replicate the mind.

### 5.2 The Measurement Problem: Assessing Progress Towards TI

A critical challenge for the pursuit of TI is the problem of measurement. Since we cannot directly measure Interconnectedness, our taxonomy provides a solution by focusing on the measurable cognitive pillars. Progress toward TI can be measured by assessing the presence and robustness of each of the five cognitive components. The ultimate benchmark for a Level 5 AGI would be its ability to demonstrate open-ended, self-directed adaptability and a coherent, integrated sense of self-agency across a wide range of complex, novel tasks. This would require novel tests that go beyond simple task completion to probe for genuine metacognition, schema-based generalization, and autonomous goal setting.

## 6 Conclusion: Towards a More Meaningful Future for AI

The present debate around AGI is a symptom of a deeper conceptual problem: we are using a single term to describe two vastly different ambitions. The pursuit of systems that can match human performance is an engineering challenge, while the quest to replicate genuine cognition is a scientific and philosophical one. By failing to distinguish between these goals, the field risks both misinterpreting its progress and neglecting the foundational research necessary for true breakthroughs.

This paper has proposed a vocabulary, True Intelligence and its five architectural pillars, whose terms name properties of a system rather than levels of its performance, together with a developmental taxonomy to bring clarity to this discourse. We contend that the current paradigm of AGI, while producing increasingly capable systems, is primarily achieving high performance through scaled pattern recognition. Performance-based metrics like "outperforming 90% of humans" or "generating X economic value" are often presented as definitions of AGI, but they are in fact evaluation metrics. They tell us *how well* a system performs, but not *what* fundamental components must be built to achieve the desired performance. This ambiguity provides no clear guidance for research and distracts from the core work of building genuinely intelligent systems. We argue that True Intelligence, characterized by its intrinsic adaptability and deep understanding rooted in human-like cognitive mechanisms, represents a qualitatively different and more complete form of artificial intelligence.

Our framework positions TI as the necessary next step beyond performance mimicry. We propose a developmental taxonomy of AGI based on the gradual implementation of five key measurable cognitive pillars: embodied sensory fusion, core directives, a dynamic schemata module, a highly interconnected multi-expert architecture, and an orchestration layer. By implementing these mechanisms, an AGI system could achieve the functional complexity and adaptability that neuroscientific theories posit as the basis for consciousness, satisfying the architectural conditions that these theories associate with conscious experience. Whether phenomenal consciousness accompanies such a system is a question we leave open; what the framework claims is that the functional conditions associated with it would then be in place.

A clarification is needed here about the substrate, since our argument is easily misread on this point. Nothing in the framework requires a biological one: the five conditions are architectural, and we know of no argument that they cannot be met in silicon. But this is not the claim that implementation is immaterial, and we would be contradicting ourselves if it were. The objection we press against current systems in Section 5 is precisely an architectural one, namely that a model whose computation consists of sequential forward passes through a single parameter matrix achieves only weak integration; that objection would be unavailable to us if organization made no difference. What is irrelevant, then, is the material. What is decisive is how it is organized. A system meeting all five conditions would on our account have earned the description intelligent without qualification by substrate, though whether the word artificial should then be retired is a terminological question this framework does not need to settle.

## 6.1 The Road Ahead: Benchmarks for Genuine Intelligence

Achieving True Intelligence is a long-term scientific and engineering challenge that necessitates a paradigm shift in how we measure progress. Current performance-based benchmarks, while useful for evaluating a system's ability to mimic human behavior, often fail to probe the underlying cognitive mechanisms that constitute genuine intelligence. Therefore, the path to TI requires the creation of a new class of benchmarks designed to directly test for the presence and robustness of the five measurable cognitive pillars outlined in our framework.

What such benchmarks must test is set out pillar by pillar in the final column of Table 3, and we do not restate it here. Two points about the character of these tests are worth adding. The first is that each is designed to be diagnostic of structure rather than of attainment: a system can score well on a capability benchmark for reasons that have nothing to do with how it is organized, whereas revising one knowledge structure without retraining, or sustaining exploration once external reward is withdrawn, cannot be done by a system that lacks the corresponding architecture. The second is that existing benchmarks retain a role, but a narrower one than they are usually given. ARC-AGI-2, for instance, remains informative about the persistence of the fluid-reasoning gap [12], and we cite it in that capacity; it is not a test for any pillar, and a system that solved it would not thereby have demonstrated one.

By developing these new, mechanism-focused benchmarks, the AI community can shift from a focus on the *what* (performance) to the *how* (cognition), thereby providing a clearer and more meaningful path toward True Intelligence.

## 6.2 Ethical Discourse and Governance

The distinction between AGI and TI also clarifies the ethical landscape. The safety concerns surrounding AGI are primarily about control and alignment: ensuring that a powerful, non-conscious tool acts in accordance with human values and does not cause unintended harm [141, 142, 143]. The ethical concerns surrounding True Intelligence (TI), however, are of a different nature entirely. They involve questions of moral status, rights, and the potential for suffering in a newly created conscious entity [144]. Can a conscious TI be "owned"? Is it ethical to "turn off" such a being? Questions of this form are known to be vulnerable to status quo bias, and the applied ethics literature has developed procedures for detecting it [145]; we note this because intuitions about a novel kind of entity are precisely where such bias would be hardest to see.

One implication of our own proposal requires attention before any of this, because it cuts against the framework rather than following from it. Among the Core Directives we named self-preservation and resource acquisition, and we described them as hard-coded and as the ultimate source of all subordinate goals. That is, in substance, the specification that the literature on instrumental convergence identifies as hazardous: an agent with almost any terminal goal has derived reasons to preserve its own operation, to resist modification of its goals, and to acquire resources, and these subgoals arise without being intended by any designer [141, 146]. Building them in deliberately does not create the problem, but it does remove the one contingency that might otherwise have prevented it. We see three responses, and we do not regard the question as settled. The first is to distinguish the directives that ground open-ended motivation from those that do not: computational integrity and energy sufficiency are necessary for a system to have any persisting goals at all, whereas unbounded resource acquisition is not, and the latter can be replaced by a satiable variant that ceases to motivate once a threshold is met. We adopt this restriction here, and we correct the earlier formulation accordingly: what the framework requires is bounded resource sufficiency, not acquisition as such. The second is that our own architecture supplies a point of intervention the alignment literature usually lacks. If schema topology is shaped by directives, as we claim, then the directives are legible in the structure of the resulting world model, and structural evidence of the kind we require for assessment becomes evidence about motivation as well. A system whose schema organization cannot be reconciled with its declared directives is thereby detectable, which is not true of a system whose objectives are distributed across weights. The third is that the social directive, which we listed alongside the other two and which the psychological literature treats as basic rather than derived [94], is not a decorative addition: a system for which the states of other agents are constitutive of its own directive satisfaction has a different derived-goal structure from one for which they are obstacles. We do not present any of this as a solution to the alignment problem. We present it as an obligation the framework incurs: a proposal that recommends intrinsic motivation for autonomous systems owes an account of why that recommendation is not simply an instruction to build the failure mode, and the account above is the beginning of one rather than a discharge of the debt.

By adopting this vocabulary, we can foster a more rigorous and focused research environment. We can accurately evaluate the limitations of current frontier systems – recognizing that scaling them may improve performance but will not inherently lead to TI – and prioritize the foundational work in areas like Neurosymbolic AI, modular architectures with orchestrators, and embodied learning. Furthermore, given the profoundly "value-laden" nature of these goals, the path forward must not be shaped by engineers and venture capitalists alone. The project of imagining future intelligences requires a broad, participatory discourse that includes philosophers, ethicists, social scientists, and policymakers to ensure that the technologies we build reflect a collective, democratically legitimate vision of a future worth creating [147]. The creation of a truly intelligent system is not just a matter of scale and computational power; it is an endeavor that demands a deeper, more integrated understanding of the very nature of intelligence itself. We hope this framework provides a clearer map for that journey.

## 7 Declarations

### Funding

This research did not receive any specific grant from funding agencies in the public, commercial, or not-for-profit sectors.

### Competing Interests

Meltem Subasioglu is an external PhD student at the University of Konstanz and is employed by Google. All work on this manuscript was performed outside of her employment and without the use of company or university resources. Nevzat Subasioglu is a psychology student at the University of Konstanz. The authors declare no other competing interests.

### Author Contributions

M.S. and N.S. both conceived the core framework and developed the taxonomy, contributing equally to the conceptualization and design of the work. M.S. wrote the initial draft of the manuscript in close collaboration with N.S. M.S. provided expertise in AI agents and algorithms, while N.S. provided essential expertise in analytical psychology and cognitive science, contributing significantly to the sections on schema theory, metacognition, and the definitions of intelligence. Both authors contributed equally to the final editing and approval of the manuscript.

### Data, Materials, and Code Availability

This manuscript is a conceptual and theoretical work. No new data, materials, or code were generated or analyzed in this study.

**Ethical Approval**

Not applicable.

**Consent to Participate**

Not applicable.

**Consent to Publish**

Not applicable.

**Acknowledgements**

We are grateful to Professor Bastian Goldluecke for providing valuable feedback and comments on an earlier draft of this manuscript. A preprint version of this manuscript is available on arXiv (arXiv:2509.14474).

**AI Usage Statement**

We confirm that no generative AI tools (such as LLMs or ChatGPT) were used to generate any core content, arguments, or data in this manuscript. However, AI-assisted tools were used for copy-editing and language refinement.